\documentclass[a4paper,fleqn,final]{cas-sc}
\usepackage[authoryear]{natbib}
\usepackage[T1]{fontenc}
\usepackage{booktabs}
\usepackage{array}
\usepackage{graphicx}
\usepackage[figuresright]{rotating}
\usepackage{subfigure}

\usepackage{bm}
\usepackage{amssymb}
\usepackage{setspace}
\usepackage{mathtools}
\usepackage{color}
\usepackage{amsmath}
\usepackage{epsfig} %% for loading postscript figures
\usepackage{multirow}
\renewcommand{\arraystretch}{1.6}

\def\ss{\scriptstyle}
\newcommand\eq{Eq.\thinspace}

\newcommand\Sec{Sec.\thinspace}

\newcommand\Fig{Fig.\thinspace}

\newcommand\T{{\ss {\rm T}}}

\newcommand\arr{\setlength\arraycolsep{2pt}\begin{array}{rl}}
\newcommand\ea{\end{array}}
\newcommand\beq{\begin{equation}}
\newcommand\eeq{\end{equation}}
\newcommand\beqa{\setlength\arraycolsep{2pt}\begin{eqnarray}}
\newcommand\eeqa{\end{eqnarray}}

\def\l#1{\label{eq:{#1}}}
\def\refe#1{(\ref{eq:{#1}})}
\newcommand\arcl{\setlength\arraycolsep{1.5pt}\begin{array}{rclrclrclrclrclrcl}}

\def\subsize{\@setsize\subsize{11pt}\xipt\@xipt}

\def\tsc#1{\csdef{#1}{\textsc{\lowercase{#1}}\xspace}}
\tsc{WGM}
\tsc{QE}
\tsc{EP}
\begin{document}
\let\WriteBookmarks\relax
\def\floatpagepagefraction{1}
\def\textpagefraction{.001}

\shorttitle{Multi-Fidelity Convolutional Autoencoder--Transfer Learning Framework}    

\shortauthors{S. Kapuria and Abhishek}  

\title [mode = title]{A Multi-Fidelity Convolutional Autoencoder--Transfer Learning Framework for Guided-Wave-Based Damage Diagnosis Using Large Simulated and Limited Experimental Datasets}  




\author[1]{Santosh Kapuria}
\cormark[1]
\ead{kapuria@am.iitd.ac.in}
\credit{Conceptualization, Investigation, Methodology, Software, Validation, Funding acquisition, Resources, Supervision, Writing - review and editing}
\orcidauthor{0000-0003-1172-1279}{Santosh Kapuria}
\author[1]{Abhishek}
\ead{abhisheksainiv2121@gmail.com}
\credit{Conceptualization, Formal analysis, Investigation, Methodology, Software, Validation, Writing - original draft}
%\orcidauthor{0000-0001-6634-3637}{Richa Kumari}
\affiliation[1]{addressline={Department of Applied Mechanics},
             organization={Indian Institute of Technology Delhi},
             city={New Delhi},
             postcode={110016},
             state={Delhi},
             country={India}}

\cortext[1]{Corresponding author}

\begin{abstract}
Guided wave-based structural health monitoring (GWSHM) with onboard transducers offers significant potential for the early diagnosis of damage in engineering structures. However, the practical deployment of deep learning models is often hindered by the limited availability of labelled experimental data and the high computational cost of generating large-scale high-fidelity simulation datasets. This study presents a multifidelity transfer learning framework that integrates lightweight physics-based simulations, convolutional autoencoder (CAE)-based deep feature learning, a feed-forward neural network, and limited experimental measurements for accurate damage localisation and sizing in plate-like structures instrumented with piezoelectric transducers. A computationally efficient one-dimensional time-domain spectral element model is employed to generate a large synthetic dataset for pretraining, while transfer learning adapts the model to experimental domains using only a small amount of labelled data. The CAE-based transfer learning framework significantly outperforms its CNN-based counterpart in damage localisation accuracy. The model achieves excellent predictive performance with $R^2$ scores exceeding 0.93 for damage localisation and 0.99 for damage sizing. Its generalisation capability is demonstrated on previously unseen data, showing high prediction accuracy for damage scenarios not represented during pretraining or fine-tuning. The results establish the proposed framework as an accurate, computationally efficient, and practically viable solution for real-world GWSHM applications.
\end{abstract}

\begin{keywords}
Structural health monitoring \sep Guided waves \sep Limited experimental data\sep Convolutional autoencoder\sep Transfer learning\sep Multifidelty deep learning 
\end{keywords}
\maketitle
\thispagestyle{empty}
\pagestyle{plain}
%%--

%------------------------------------------------------------------------------------------------------------
% SECTION 2:
%------------------------------------------------------------------------------------------------------------

%\date{}

\section{Introduction}
Structural health monitoring (SHM) has emerged as a critical research area aimed at enhancing the safety, reliability, and longevity of engineering structures while preventing catastrophic failures, minimising downtime and reducing life-cycle costs through the early detection and characterisation of damage~\citep{farrar2007introduction}. Among the various SHM techniques, guided wave-based monitoring has attracted particular attention due to its ability to inspect large structural regions using a relatively small number of sensors and to detect even the smallest and most subtle surface and subsurface defects in thin-walled structures~\citep{SuYe2009,rose2014ultrasonic,cawley2024guided}. It has been widely adopted in a variety of applications, including aircraft fuselages and wings, space structures,  ship hulls, pipelines, process vessels, and wind turbine blades~\citep{DaltonEtal2001, GiurgiutiuEtal2011, WangEtal2018, HuEtal2020, HuanEtal2020, kannusamy2025lamb, dadashbaki2025autonomous}. This widespread adoption has been further facilitated by the convenient, efficient, and cost-effective generation and sensing of guided waves through surface-bonded or embedded piezoelectric wafer active sensors (PWAS)~\citep{giurgiutiu2014structural}.

The fundamental principle of guided wave-based SHM (GWSHM) relies on analysing the interaction between propagating waves and structural discontinuities caused by defects. As guided Lamb waves encounter defects, their propagation characteristics, including amplitude, phase, and time of flight (ToF), are altered, and extra wave modes emerge due to scattering, reflection, and mode conversion. These changes provide valuable information regarding the presence, location, and severity of damage. A commonly employed strategy is the baseline approach, which utilises deviations from the baseline signals of the healthy state of structures to identify and locate damage~\citep{michaels2007guided,dawson2016isolation,kudela2018structural,HuEtal2020,HuanEtal2020}. However, owing to the multimodal and dispersive nature of Lamb waves, as well as the complex geometry of real-world structural components, the measured signals are often distorted and contaminated by multiple wave modes and boundary reflections~\citep{zhang2022effective,azad2025integrated}. Consequently, appropriate signal processing algorithms~\citep{JanaKapuria2026,song2026physics} are required to accurately isolate the desired wave packets and extract damage-sensitive features. This task is often challenging and, in some cases, even intractable, thereby limiting the accuracy of these physics-based damage detection techniques.

Baseline-dependent methods also suffer from limitations under varying environmental and operational conditions, owing to the sensitivity of baseline signals to such variations, which may result in false alarms~\citep{dodson2013thermal,rezazadeh2025systematic}. To address these challenges, several baseline-free techniques have been proposed~\citep{zhu2024appraisal}, among which the refined time-reversal method~\citep{Agrahari} has shown considerable promise under varying temperature conditions~\citep{Kannusamy_2020,sharma2021time,sharma2023baseline}.
However, these methods also require the desired single or multiple Lamb wave modes to be isolated from the forward response so that undesirable overlapping modes do not distort the reconstructed signals obtained after the time-reversal process. %%This step also requires sophisticated signal processing, for which the available techniques are often insufficient, particularly in real-world applications~\citep{JanaKapuria2026}.
This signal-processing step also faces challenges similar to those discussed above for baseline-based methods~\citep{kapuria2025best}.

Because of the aforementioned limitations of physics-based GWSHM methods, there has been a growing interest in data-driven approaches, particularly machine learning (ML) and deep learning (DL) techniques, which can automatically extract representative features from raw or minimally processed signals~\citep{yang2023review}. Rather than relying on predefined features such as the ToF, these methods utilise the full waveform information to establish nonlinear mappings between wave responses and damage parameters, thereby enabling not only damage detection but also its localisation and characterisation.

%Autoencoders, with their encoder--decoder architecture, are especially effective for unsupervised feature extraction, dimensionality reduction, and signal reconstruction, enabling compact latent representations that preserve damage-sensitive information. 
Among various DL architectures, convolutional neural networks (CNNs) have demonstrated remarkable potential for GWSHM due to their inherent ability to learn hierarchical spatial and temporal features from complex wavefield data. Early applications of CNNs to GWSHM transformed Lamb wave signals into two-dimensional (2D) image representations, for example, through time--frequency analysis~\citep{ewald2019deepshm,liu2020deep}. \citet{miorelli2021defect} employed a 2D-CNN to detect and localise defects in aluminium plates using damage amplitude spectrograms as input. Similarly, 2D-CNNs have been used to detect internal delaminations in composite structures from time--frequency images of Lamb wave signals obtained using the continuous wavelet transform~\citep{wu2021lamb}.
Subsequently, one-dimensional (1D) CNNs gained attention due to their ability to directly process raw signal data. For instance, \citet{rai2021lamb} employed a 1D-CNN to classify healthy and  damaged states using raw Lamb wave signals for training and prediction. Likewise, \citet{zhang2021damage} utilised 1D-CNNs to predict damage locations based on a time-varying damage index derived from the discrete Fourier transform of Lamb wave signals. More recently, 1D-CNN architectures have been integrated with transfer learning to achieve the dual objectives of damage localisation and severity assessment in composite plates using raw Lamb wave data~\citep{azad2025integrated}. Another emerging research direction involves incorporating explainability techniques into CNN-based damage detection frameworks to improve model interpretability and foster user trust~\citep{pandey2022explainable,lomazzi2023explainability}.

Collectively, these studies demonstrate the growing capability of deep CNNs to enhance the precision and efficiency of Lamb wave-based damage assessment using raw GW data. Nevertheless, the performance of data-driven methods in field applications is strongly dependent on the quantity and quality of the available training data. Since these models must ultimately make predictions based on measurements acquired from actual structures operating under real-world conditions, the training data should ideally be drawn from the same underlying distributions to ensure good generalizability and transferability~\citep{vapnik1999overview}. However, experimental data from real structures are often expensive and difficult to obtain, posing a major challenge to the successful deployment of ML/DL models in SHM applications. In contrast, simulation data can be generated more easily and at a significantly lower cost, motivating many researchers to use synthetic datasets for model training and validation~\citep{sbarufatti2014numerically,de2015application}. Nevertheless, simulated data may not adequately capture the variability present in experimental measurements due to systematic modelling errors and the presence of measurement noise and other uncertainties. As a result, models trained primarily on simulation data may yield inaccurate predictions when applied to experimental data, particularly for damage localisation and characterisation.

The aforementioned limitations highlight the need for practical approaches that leverage large simulation datasets alongside limited experimental data to train ML models. Model updating techniques iteratively adjust model parameters to improve the agreement between simulated and measured responses~\citep{agathos2021crack}.  However, they require all significant sources of variability in the experimental data to be adequately represented in the model. To address this challenge, \citet{olleak2020calibration} proposed a Gaussian process regression-based framework for calibrating simulation data using limited experimental observations in the context of selective laser melting. Similarly, \citet{zhang2022effective} employed the metric learning for kernel regression (MLKR) approach to perform supervised learning on combined simulation and experimental datasets through bias correction and noise filtering. More recently, \citet{nerlikar2024physics} developed a convolutional autoencoder (CAE)-based framework to obtain data closely resembling experimental measurements from simulated GW signals.  However, their framework requires comparable amounts of simulation and measurement data and, therefore, does not address the practical challenge of the limited availability of experimental data for training.

During the past decade, transfer learning (TL) has emerged as a powerful strategy for adapting an ML model originally trained on a source domain with abundant data to solve a different but related problem in a target domain where only limited data are available, by leveraging the knowledge acquired from the source domain~\citep{weiss2016survey,zhuang2020comprehensive}. This approach enables the model to learn domain-specific features of the target domain rather than merely reducing the discrepancy between the source and target data distributions. 
Despite its growing popularity, the application of TL for transferring knowledge from physics-based simulations to real-world domains remains relatively unexplored. \citet{lin2022dynamics} presented a simulation-to-experiment TL framework based on two parallel 1D-CNNs for vibration-based damage detection in a simple structure, namely a simply supported beam. Similarly, \citet{BaoEtal2023} implemented a TL framework comprising a 1D-CNN followed by fully connected layers for damage scenario classification in bolted steel connections using abundant simulation data and limited experimental measurements of acceleration responses. These studies highlight the potential of combining physics-based simulations with data-driven learning to improve model generalisation while substantially reducing the requirement for experimental data.
However, to the best of the authors' knowledge, a simulation-to-real-world TL framework for GWSHM capable of both damage localisation and severity estimation has not yet been reported.

Another aspect that warrants attention is the computational effort involved in generating simulation data, not only from the perspective of reducing computation cost and runtime but also in terms of energy efficiency and sustainability~\citep{oishi2022sustainable}. Unlike low-frequency vibration analysis, high-frequency wave propagation analysis requires fine spatial and temporal discretisation, leading to a substantial increase in computational cost and runtime. Furthermore, to realistically model PWAS transducers and various damage scenarios (e.g., notches and corrosion), continuum-based multiphysics finite element (FE) models are commonly employed~\citep{zhang2021damage}, further increasing the computational burden.
To alleviate this challenge, it is important to adopt lightweight yet accurate simulation models for Lamb wave generation, propagation, and sensing. Significant advances have been made in developing structural theory-based multiphysics FE formulations for beam-, plate-, and shell-type structures, achieving several-fold reductions in problem size compared with continuum-based FE models~\citep{KapuriaHagedorn2007,kapuria2021coupled,velasquez2022finite,najd2025variable}. To further improve computational efficiency, nonconventional FE techniques such as the time-domain spectral element (TDSE) method~\citep{rekatsinas2017time,JAIN2024,jain2025efficient} and the local-domain wave packet-enriched finite element method~\citep{kapuria2020wave} have been developed, yielding up to an order-of-magnitude reduction in computational time relative to conventional FE method. Despite these advances in computationally efficient and accurate simulation tools, their effectiveness within a deep transfer learning framework for knowledge transfer from the simulation domain to the measurement domain has not yet been investigated.

This work presents a multifidelity convolutional autoencoder-based transfer learning framework for GWSHM of plate-like structures instrumented with PWAS transducers, combining a large low-fidelity simulation dataset with a limited amount of high-fidelity experimental data to address the inverse problem of damage localisation and severity assessment in real-world applications. The framework directly processes normalised raw Lamb wave signals without requiring complex signal preprocessing or feature engineering. As a further novelty, the proposed framework leverages the computational efficiency of a recently developed one-dimensional TDSE (1D-TDSE) model based on the zigzag theory~\citep{JAIN2024} for smart laminated panels, which is capable of physically modelling patch-type PWAS actuators and sensors as well as asymmetric notch damage in plates, to generate the large synthetic dataset required for network pretraining. First, the CAE is pretrained using the same 1D-TDSE-generated Lamb wave signals as both inputs and outputs. This process enables the network to learn a low-dimensional latent representation that captures the essential characteristics of the Lamb wave signals, while accurately reconstructing the original responses. The resulting latent-space features are subsequently used to pretrain a feed-forward neural network (FFNN) for predicting damage location and severity. During transfer learning, the encoder of the pretrained CAE is kept frozen, whereas the decoder and FFNN are fine-tuned using a limited set of labelled experimental data collected from an aluminium plate instrumented with lead zirconate titanate (PZT) transducers and subjected to block-mass damage under Hann-windowed tone-burst excitation. The performance of the proposed framework is first assessed using a large 1D-TDSE dataset and a relatively small set of high-fidelity 2D FE simulation results corresponding to a single excitation frequency, both representing notch-type damage. Subsequently, the framework is validated using experimental measurements obtained with block-mass damage at multiple excitation frequencies, thereby demonstrating knowledge transfer across different damage scenarios. Finally, the generalisation capability of the proposed framework is established through testing on previously unseen data, which shows high predictive accuracy for both damage localisation and size estimation.

%By combining the diversity and scalability of numerical simulations with the physical realism of experimental observations, the proposed hybrid framework effectively mitigates the challenge of limited labelled experimental data. This strategy enables the learned models to generalize well across different damage scenarios while maintaining high predictive accuracy. The study demonstrates that transfer learning, coupled with deep feature extraction, provides a robust and practical solution for Lamb wave-based SHM, achieving reliable notch localization and depth estimation and highlighting its potential for deployment in real-world structural integrity assessment applications.

\section{Overview of Methodology}
This section presents the multi-fidelity physics-embedded data-driven formulation framework developed to address the inverse problem of damage localisation and sizing in Lamb wave-based SHM. The overall objective of the proposed ML framework is to predict the location and size of defects in a thin-walled structure from real-time measurements of guided wave signals, using a large set of physics-based model data and a limited set of experimental/real-world data for training. The proposed model adopts a convolutional autoencoder-driven feed-forward neural network with transfer learning for autonomous assessment of structural health, and is herein named the CAE-FFNN-TL model. The flow chart of the CAE-FFNN-TL model framework is shown in Fig.~\ref{fig:CAE-FFNN-TL-flowchart}. Before discussing the details of the data generation, network architecture  and training strategies, an overview of the proposed methodology is presented first, which comprises the following three components: 
\begin{figure}[!htp]
    \centering
 \subfigure[Pretraining and testing framework with large low-fidelity simulation data]{      \includegraphics[scale=0.63, trim = 0.5in 1.8in 0in 1in, keepaspectratio]            {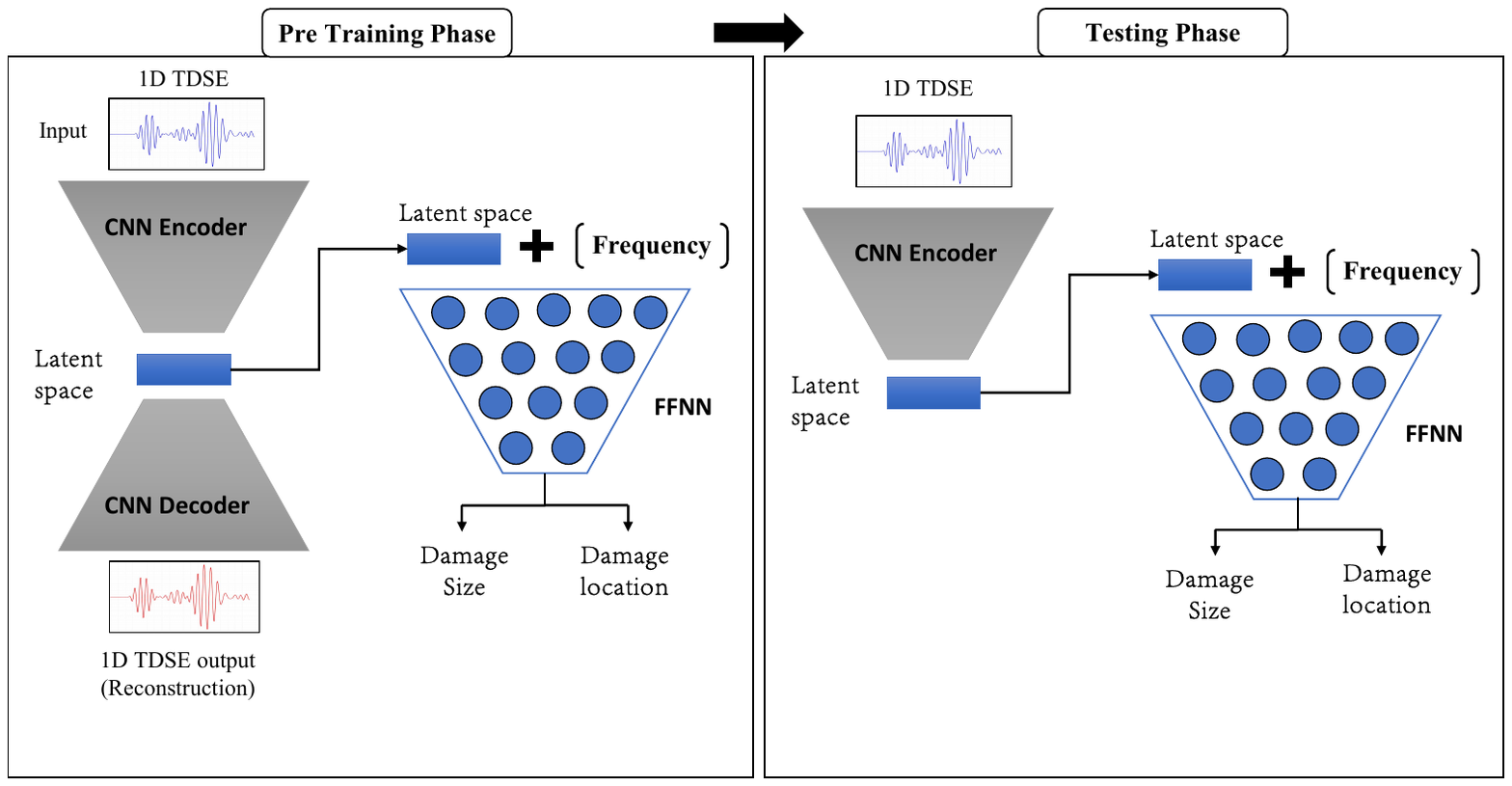}}
      \label{fig:CAE-FFNN-TL-flowchart(a)}
  \subfigure[Transfer learning with limited high-fidelity experimental/real world data]    {\includegraphics[scale=0.63, trim = 0.5in 1.6in 0in 1.2in, keepaspectratio] {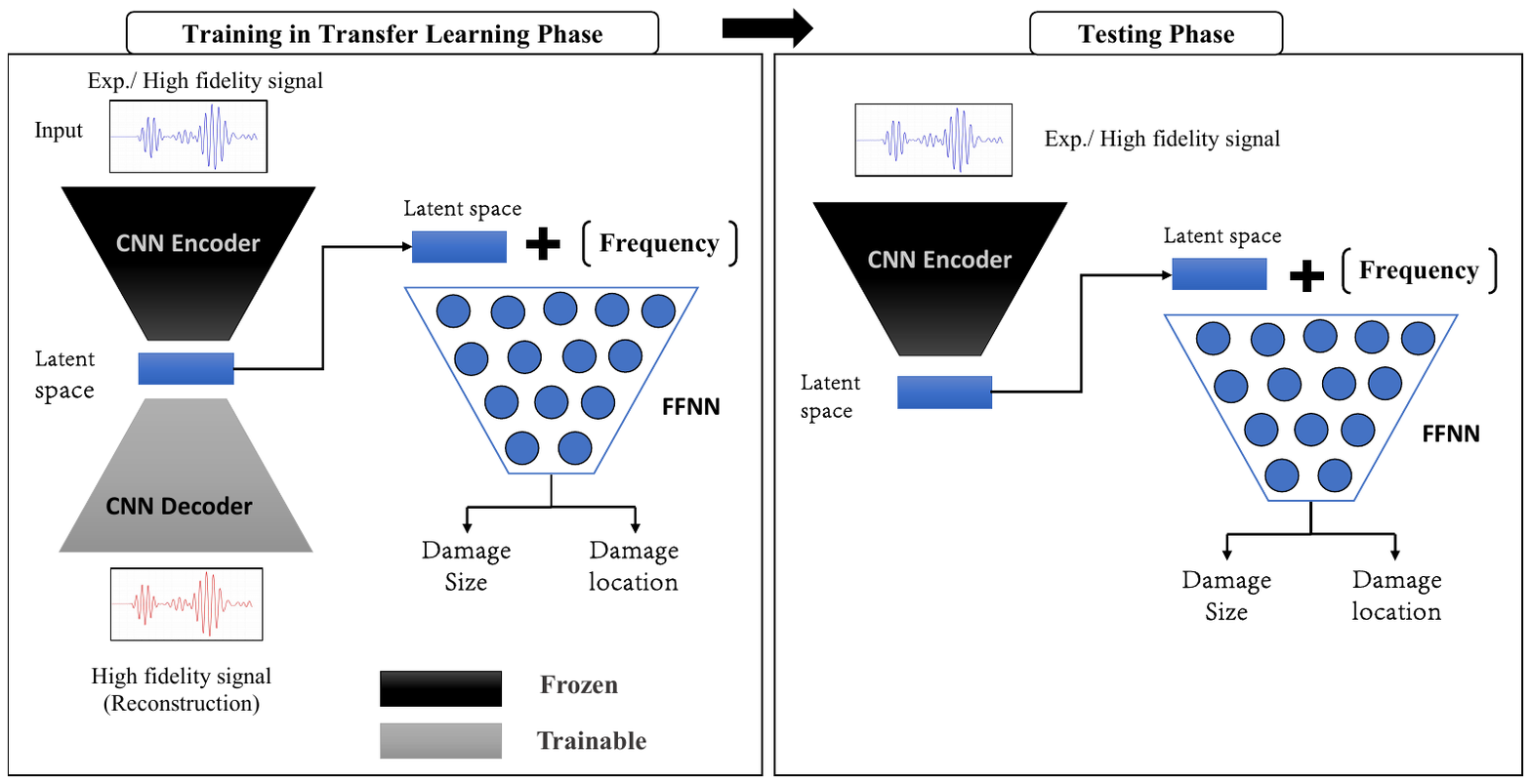}}
     \label{fig:CAE-FFNN-TL-flowchart(b)}
    \caption{High-level flow chart for training and testing phases of the proposed CAE-FFNN-TL model for guided wave-based SHM}
    \label{fig:CAE-FFNN-TL-flowchart}
\end{figure}
\begin{enumerate}
\item [(i)]The first step is to generate guided wave propagation response data using an efficient physics-based model. Guided-wave-based SHM techniques typically employ high-frequency excitation signals (50 kHz-1 MHz) to detect small defects. The high frequency makes the simulation of wave-propagation response using the conventional FE models available in commercial software very time-consuming. To overcome this challenge, a computationally efficient 1D-TDSE model based on the third-order zigzag theory developed by \citet{JAIN2024,jain2025efficient} is used here to simulate the generation, propagation, and sensing of Lamb waves in the host structure with and without damage, through integrated PWAS transducers.
    \item [(ii)]In the second step, a CAE network is pretrained with a large dataset of normalised raw Lamb wave signals obtained from 1D TDSE simulations (after being augmented with noise) as both input and output. Its encoder-decoder structure with a latent space enables highly low-dimensional, robust feature extraction from the high-dimensional Lamb wave time-series data. Upon completion of CAE pretraining, latent representations are extracted for each input signal and concatenated with their corresponding excitation frequency values. These combined features are used as inputs to an FFNN, which is trained to predict damage size and location from the time-history signal.
    \item [(iii)]In the third step, the pretrained CAE-FFNN model is fine-tuned on a limited dataset of experimentally acquired sensor output (or high-fidelity simulation data) through transfer learning to make the model amenable to the target real-world domain.
\end{enumerate} 

Brief descriptions of the model's components are provided below.

\section{1D TDSE Model for Simulating Lamb Wave Actuation and Sensing}
\label{sec:1D TDSE Formulation}
\begin{figure}[!htp]
\centering
\includegraphics[scale=0.6, trim = 0in 1.8in 0in 1.4in, keepaspectratio]{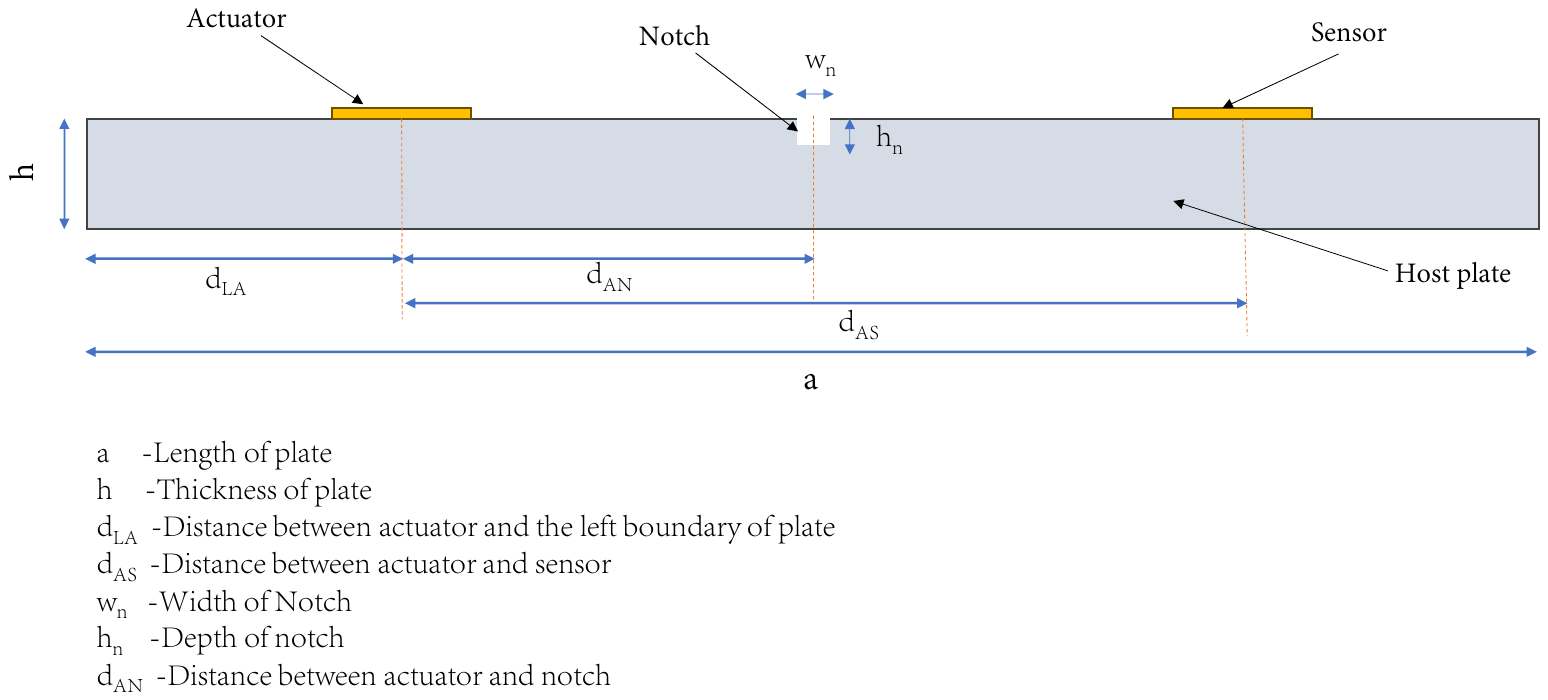}
\caption{Schematic diagram of the geometry of host plate integrated with surface-bonded PWAS transducers and featuring a notch-type damage} \label{Fig: TDSE host plate}
\end{figure}

Kapuria and coworkers~\citep{JAIN2024,jain2025efficient} have devised a computationally efficient 1D TDSE model based on the efficient layerwise third-order zigzag theory (ZIGT), demonstrating fast and accurate predictions of guided wave propagation in structures such as beams and panels (made of isotropic or laminated composite materials) integrated with PWAS patches. The formulation physically models the piezoelectric transducers of finite thickness and incorporates two-way electromechanical coupling, thereby enabling accurate prediction of the induced sensor potential due to wave propagation in the host structure. Figure~\ref{Fig: TDSE host plate} shows a typical actuator-host plate-sensor configuration with a notch-like damage. Although the present study deals with the SHM of single-layer metallic plates, the introduction of piezoelectric patch transducers makes these sections a layered structure. Using commercially available FE analysis packages, such a configuration would typically be modelled using continuum-based two-dimensional (2D) elements. However, this would be highly time-consuming for simulating high-frequency wave propagation, making it unsuitable for generating a large dataset. The present zigzag theory-based 1D TDSE model achieves a significant reduction in the computation time by reducing the number of degrees of freedom (DoFs) (i) in the thickness direction due to the kinematic approximation and (ii) in the longitudinal direction due to the use of spectral interpolation functions. The asymmetric notch can also be conventionally modelled using the zigzag model by treating the portion below the notch as a separate layer. The model is briefly described below for clarity and completeness. An interested reader may refer to~\citep{JAIN2024} for further details. 

\subsection{Electric Potential and Displacement Field Approximations}
Piezoelectric extension-mode actuation yields a nearly quadratic through-thickness electric potential in piezo layers. The laminate thickness is partitioned into $n_{\phi}$ piezo interfaces and electrode surfaces that are equipotential. The through-thickness variation of electric potential $\phi$ is considered to comprise a piecewise-linear part (capturing surface/interface potentials) plus a quadratic part:
\begin{equation}
\phi(x,z,t)=\sum_{j=1}^{n_\phi}\Psi_\phi^{\,j}(z)\,\phi^{j}(t)\;+\;\sum_{l=1}^{n_\phi-1}\Psi_c^{\,l}(z)\,\phi_c^{\,l}(x,t),
\label{eq:phi_approx}
\end{equation}
where $\Psi_\phi^{\,j}(z)$ are piecewise-linear functions in $z$, while $\Psi_c^{\,l}(z)$ are quadratic functions spanned over each sublayer. This separation of $\phi^j$ and $\phi^l$ enables accurate satisfaction of equipotential conditions at electrode surfaces, while resulting in a significant reduction in the number of electrical DoFs. 

Thickness-polarized piezoelectric layers induce $\varepsilon_z \approx \bar d_{33}^{(k)} E_z$. Integrating this relation through the thickness yields the transverse displacement $w$ as
\begin{equation}
w(x,z,t)
=
w_0(x,t)
-\sum_{j=1}^{n_\phi}\phi^{j}(t)\,\bar\Psi_\phi^{\,j}(z)
-\sum_{l=1}^{n_\phi-1}\phi_c^{\,l}(x,t)\,\bar\Psi_c^{\,l}(z),
\label{eq:w_approx}
\end{equation}
where $w_0$ denotes the reference-plane deflection and
$\bar\Psi_\phi^{\,j}$ and $\bar\Psi_c^{\,l}$ are thickness integrals of
$\Psi_\phi^{\,j}$ and $\Psi_c^{\,l}$ weighted by $\bar d_{33}^{(k)}$
over the corresponding sublayers.
\begin{figure}
\centering
\includegraphics[scale=0.4, trim = 0in 2.5in 0in 2.0in, keepaspectratio]{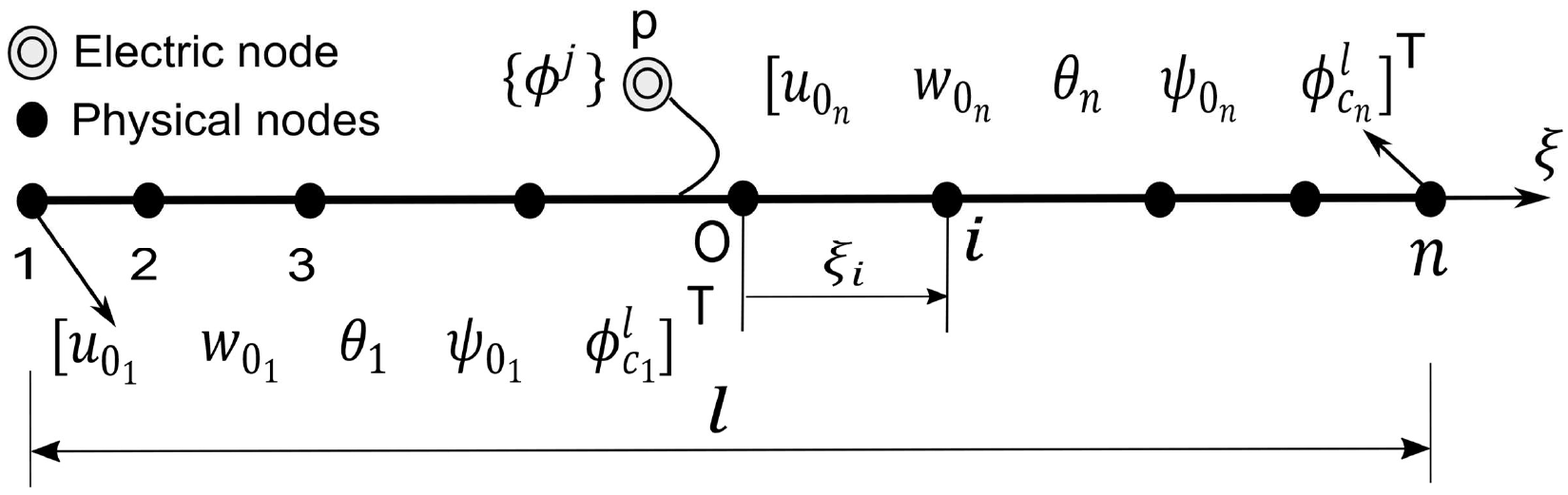}
\caption{Spectral strip element based on the ZIGT, showing $n$ physical nodes with four mechanical DoFs and $n_\phi-1)$ internal electric potential DoFs  per node and an electric node (p) with $n_\phi$ surface electric potential DoFs}\label{fig:1D TDSE Spectral strip}
\end{figure}
A layerwise third-order zigzag kinematics is adopted for the axial displacement $u$, which is given by 
\begin{equation}
u(x,z,t) = u_0(x,t) - z\,w_{0,x}(x,t) + R^{k}(z)\,\psi_0(x,t),
\label{eq:u_reduced}
\end{equation}
where $R^{k}(z)$ is a layerwise cubic function determined by the
layer thicknesses, positions, and material properties. The advantage of using this kinematic model is that it allows for the slope discontinuities at the layer interfaces of the laminate, yet is ultimately expressed in terms of only three global displacement variables, independent of the number of laminae, making it computationally very efficient. Additionally, unlike other laminate theories with a comparable number of DoFs, it satisfies the continuity of transverse shear stresses at layer interfaces and  the zero transverse shear traction conditions on the plate surfaces, thereby achieving superior accuracy.

\subsection{Time-Domain Spectral Element (TDSE)}
The equations of motion of the system are solved in weak form using the extended Hamilton's principle. Based on the zigzag theory, the volume integral for the domain reduces to an integral over the $x$-direction only. The 1D domain is discretised by employing a multi-physics spectral strip element of length $l^e$ with $n$ physical nodes and one virtual electric node p (no physical coordinate). 
The mechanical DoFs $\{u_0,w_0,\theta(=w_{0,x}),\psi_0\}$ and the internal electric DoFs $\phi_c^l$ vary along $x$ and are interpolated using physical nodes, whereas the surface electric potentials $\phi^j$ are constant within an element and are assigned to the electric node. If a patch spans multiple elements, the same $\phi^j$ can be shared via node $p$ to reduce electrical DoFs~\citep{KapuriaHagedorn2007}. 

Using the natural coordinate $\xi\in [-1,1]$, $u_0,\psi_0$, and $\phi_c^l$ are interpolated with $C^0$-continuous Lagrange functions $N_i(\xi)$, while $w_0$ and $\theta$ use $C^1$-continuous generalized Hermite functions $\bar N_i(\xi)$ \citep{Pozrikidis2005,KapuriaJain2021}. Unlike the conventional FE method, the nodes in the spectral element are not equidistant, but located at the roots of the completed Gauss-Lobatto-Legendre (GLL) polynomials of $n$th order. This process makes the interpolation functions mutually orthogonal, resulting in spectral convergence [$\propto (1/n)^n$]. The interpolation can be expressed in terms of the DoFs of element $e$ as
\beq
u_0=\mathbf{N}\,\mathbf{u}_0^e,\quad \psi_0=\mathbf{N}\,\boldsymbol{\psi}_0^e,\quad
w_0=\bar{\mathbf{N}}\,\mathbf{w}_0^e,\quad
\phi_c^l=\mathbf{N}\,\boldsymbol{\phi}_c^{l,e},
\l{SE_interp}
\eeq
where \beq
\mathbf{N}=\begin{bmatrix}N_1&\cdots&N_n\end{bmatrix},\quad
\bar{\mathbf{N}}=\begin{bmatrix}\bar N_1&\cdots&\bar N_{2n}\end{bmatrix},\quad
\mathbf{w}_0^e=\begin{bmatrix}w_{0_1}&\theta_1&\cdots&w_{0_n}&\theta_n\end{bmatrix}^{\T},
\l{SE_shapes}
\eeq
with $\mathbf{u}_0^e=[u_{0_1}\ \cdots\ u_{0_n}]^{\T}$, $\boldsymbol{\psi}_0^e=[\psi_{0_1}\ \cdots\ \psi_{0_n}]^{\T}$ and $\boldsymbol{\phi}_c^{l,e}=[\phi_c^{l,1}\ \cdots\ \phi_c^{l,n}]^{\T}$. The element DoFs are arranged to form the elemental generalised displacement vector $\mathbf{U}^e$ as
\beq
\mathbf{U}^e=\begin{bmatrix}
u_{0_1}&w_{0_1}&\theta_1&\psi_{0_1}&\phi_c^{l,1}&\cdots&
u_{0_n}&w_{0_n}&\theta_n&\psi_{0_n} \quad \phi_c^{l,n} \quad \phi^j
\end{bmatrix}^{\T}.
\l{SE_Ue}
\eeq 

Using the interpolation given by \eq\refe{SE_interp}, the strain-displacement and electric field-potential relations, and the linear constitutive relations for an orthorhombic piezoelectric material, the variational integral in the extended Hamilton's principle is computed by summing up the contributions from all the $n_e$ elements as
\beq
T=\sum_{e=1}^{n_e} \delta{\mathbf{U}^e}^{\T}\left(\bar{\mathbf{M}}^e\ddot{\mathbf{U}}^e+\bar{\mathbf{K}}^e\mathbf{U}^e-\bar{\mathbf{P}}^e\right),
\l{SE_weak}
\eeq
where integrals in the element mass metrics $\bar{\mathbf{M}}^e$, stiffness matrix $\bar{\mathbf{K}}^e$, and load vector $\bar{\mathbf{P}}^e$ are evaluated using the Lobatto quadrature~\citep{Pozrikidis2005} with ($2n-1$), ($2n+1$), and ($n+1$) integration points, respectively. 

\subsection{Semi-Discrete Equations of Motion for Actuation-Sensing Mode}
The summation in \eq\refe{SE_weak} results in the formation of the global DoF vector $\mathbf{U}$, load vector $\mathbf{P}$, and matrices $\mathbf{M}$ and  $\mathbf{K}$ by an assembly process and yields the semi-discrete equations of motion for the electromechanical system as 
\beq
\mathbf{M}\,\ddot{\mathbf{U}}+\mathbf{K}\,\mathbf{U}=\mathbf{P}.
\l{global}
\eeq
With actuator potentials $\boldsymbol{\Phi}_a$ prescribed and sensor potentials eliminated through static condensation using  zero charge condition and introducing the proportional damping matrix $\mathbf{C}^{uu}$, the reduced mechanical system can be obtained as
\beq
\mathbf{M}^{uu}\,\ddot{\bar{\mathbf{U}}}+\mathbf{C}^{uu}\,\dot{\bar{\mathbf{U}}}+\bar{\mathbf{K}}^{uu}\,\bar{\mathbf{U}}
=\bar{\mathbf{P}}-\mathbf{K}^{ua}\,\boldsymbol{\Phi}_a,
\l{reduced}
\eeq
with the effective coupled stiffness 
$\bar{\mathbf{K}}^{uu}=\mathbf{K}^{uu}-\mathbf{K}^{us}\left(\mathbf{K}^{ss}\right)^{-1}\mathbf{K}^{su}$, and the unknown sensor potentials determined as 
\beq
\mathbf{\Phi_s}=-[\mathbf{K}^{ss}]^{-1} \mathbf{K}^{su}\bar{\mathbf{U}}.
\l{eig}
\eeq

Equation~\refe{eig} is solved using the Newmark-$\beta$ direct time-integration technique with $\beta=0.25$  and $\gamma=0.5$.
Because of its accurate and innovative capture of the through-thickness variations in the displacement and electric potential fields, and its high computational efficiency, the above 1D TDSE model is a good candidate for generating a large dataset of wave-propagation-induced sensor outputs to train the proposed CAE-FFNN-TL framework for SHM applications.

\section{CAE-FFNN-TL Multi-Fidelity Model for SHM}
\label{sec:First Model}
This section describes the proposed convolutional autoencoder-driven transfer learning model shown in \Fig\ref{fig:CAE-FFNN-TL-flowchart}. Brief descriptions of the key elements of the deep learning network are provided below. These include the basic structures of FFNNs, the commonly used regression loss functions, the Adam optimisation algorithm, 1D CNNs, and autoencoders. 
%\begin{figure}
%\centering
%\includegraphics[scale=0.6, trim = 0in 1.8in 0in 1in, keepaspectratio]{CAE_TL_1.pdf}
%\caption{\texttt{CAE\_FFNN\_TL} model achitecture trained on large 1D TDSE Lamb wave response} \label{Fig: CAE_FFNN_Model pretraining phase }
%\end{figure}
\subsection{1D Convolutional Neural Network}
CNNs are a powerful class of deep learning models specifically tailored for analysing data with a grid-like structure, such as 1D time series, 2D images, and 3D videos~\citep{GoodfellowEtal2016}. The 1D CNN, therefore, provides a natural choice for handling time-domain Lamb wave signals as inputs for guided-wave-based SHM. CNNs efficiently reduce the number of network parameters compared to fully connected models, making them highly effective for learning from data with complex features. It can be used for efficient, automated feature extraction. The other significant advantages of using CNNs include the ability to use raw time-series data without preprocessing or feature handcrafting, time-shift invariance, and noise robustness. CNNs learn data patterns through backpropagation using multiple building blocks, such as convolution, pooling, and fully connected layers, as discussed below. 

\subsubsection{Convolutional layer}
A convolution layer transforms an input signal into feature maps by applying learnable filters (or kernels), much smaller than the input.
During forward propagation, each kernel $K$ of size $k$ is slid across the input $I$ of size $l$ step by step, with each step referred to as a stride $S$. The feature map output at position $i$ is obtained by applying a nonlinear activation function $f(\cdot)$ on the convolution of $K$ with $I$ as
\begin{equation}
\label{Eqn:convolution}
O_i = (I * K)_i=f\left( \sum_{m=0}^{k-1} I(i - m)\,K(m) + b \right),
\end{equation}
where $b$ is the bias. The 1D convolution operation is shown in Fig.~\ref{Fig:ConvOperation}(a).
%\begin{figure}[!htp]\centering
%\includegraphics[scale=0.4, trim = 0in 3.3in 0in 3in, keepaspectratio]{Fig4_Convolution.pdf}
%\caption{1D convolution operation}\label{Fig:Convolution_operation}
%\end{figure}

In a convolutional layer, multiple kernels can be used, generating multiple feature maps, each detecting a different local temporal pattern in the input sequence. The kernel size and number, and the stride, are hyperparameters that should be determined based on the input data. Before applying a kernel to the input data, it is padded by adding extra points at its borders to control the spatial dimensions of the output feature map. If padding $P$ is applied, then the output size is given by 
\begin{equation}
\label{Eqn:convolution_output_size}
    O = \frac{l-k+2P}{S}+1.
\end{equation}

\subsubsection{Pooling layer}
\begin{figure}[!htp]
\centering
\subfigure[]{\includegraphics[scale=0.4, trim = 0in 3.3in 0in 3in, keepaspectratio]{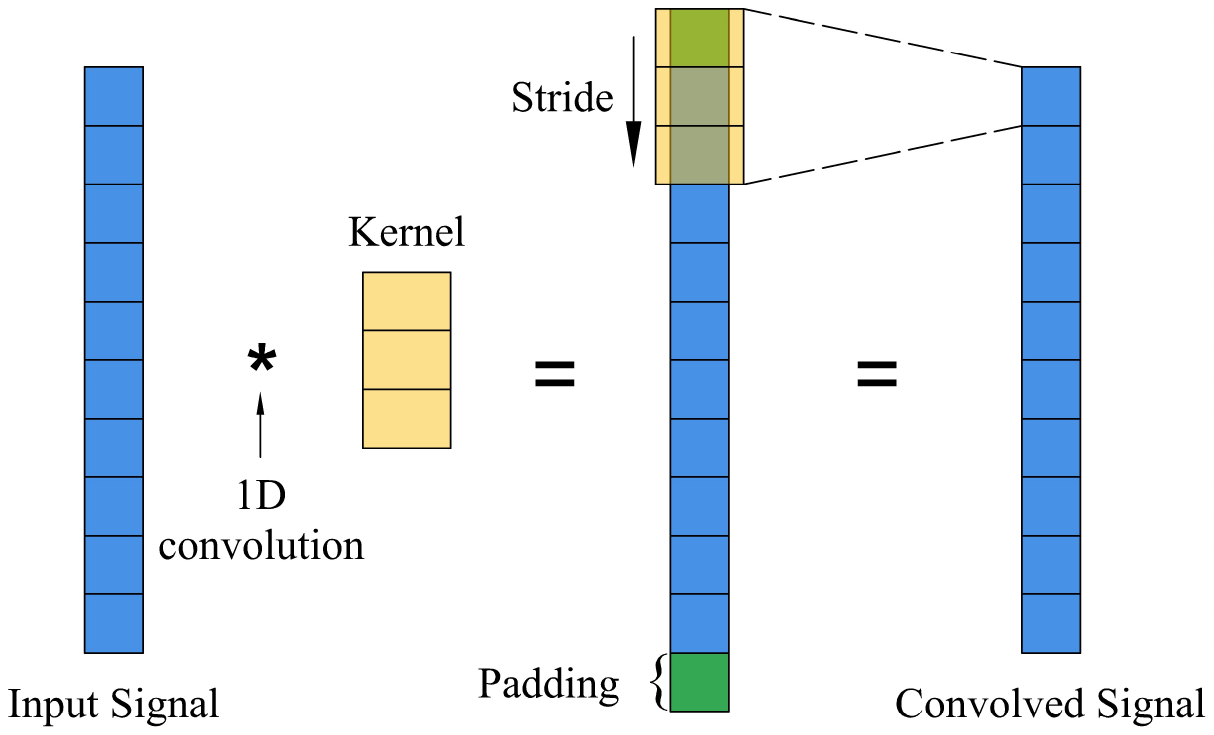}}
\subfigure[]{\includegraphics[scale=0.25, trim = 0in 2in 0in 1.0in, keepaspectratio]{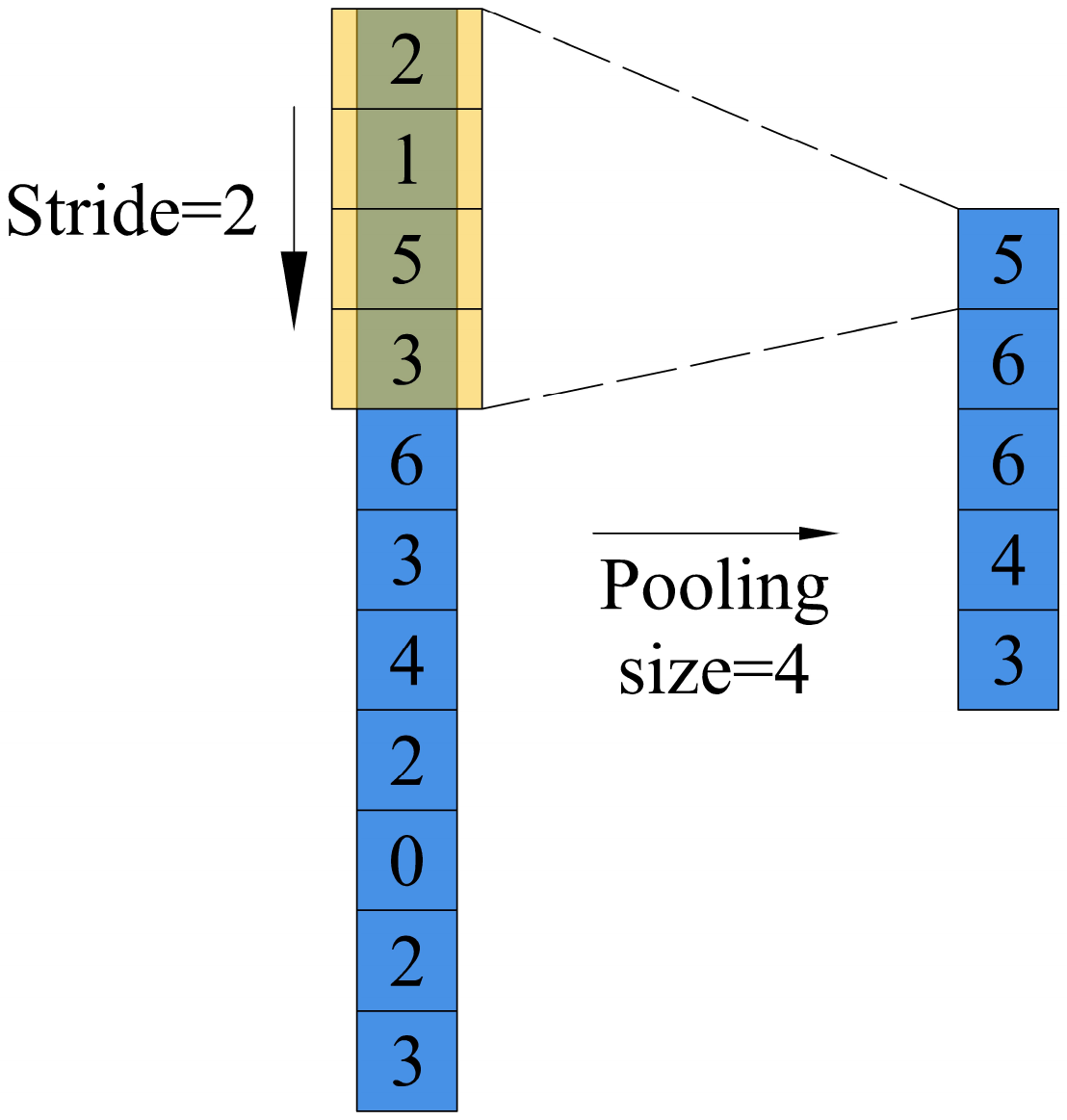}}
\caption{(a) 1D convolution operation. (b) Max-pooling operation}\label{Fig:ConvOperation}
\end{figure}

A pooling layer in neural networks decreases the dimensionality of data while preserving its essential features. This reduction is achieved by applying mathematical operations, such as max pooling or average pooling, to groups of input data, yielding a condensed representation of the original data. Pooling layers are frequently utilised to mitigate overfitting and enhance computational efficiency. A typical max-pooling operation with stride~$=2$ and pooling size~$=4$ is shown in Fig.~\ref{Fig:ConvOperation}(b).

\subsubsection{Fully-connected layer}
\begin{figure}[!htp]
\centering
\includegraphics[scale=0.35, trim = 0in 0in 0in 0.5in, keepaspectratio]{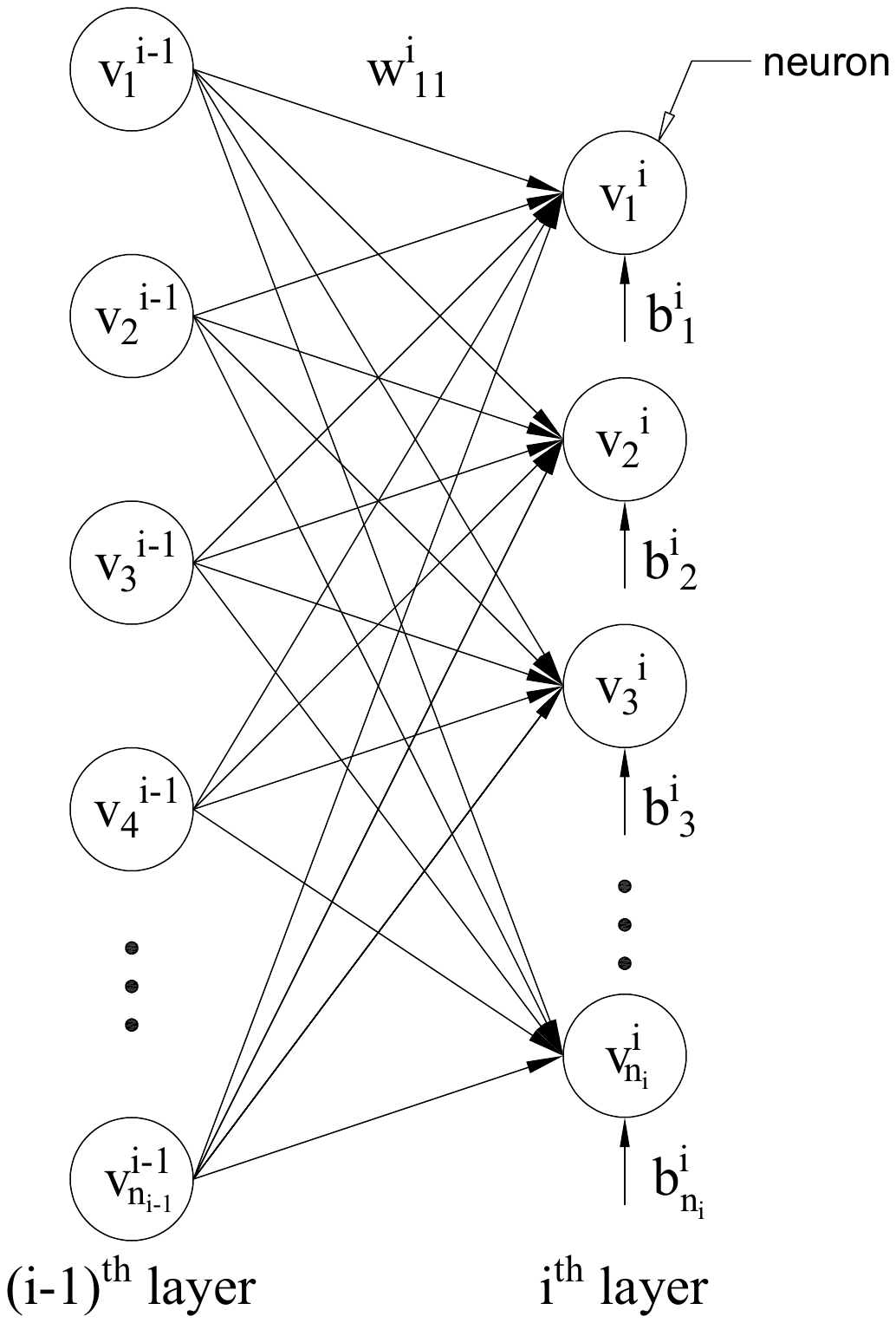}
\caption{Connection between neurons of adjacent layers in a fully-connected layer }\label{Fig:ANNmath}
\end{figure}

A fully connected layer, also known as a dense layer, is one where every neuron in the previous layer is connected to every neuron in the output layer. Dense layers are used to achieve global integration of learned features. Typically, the output of these layers is passed through an activation function $f(\cdot)$. The connection between neurons in adjacent dense layers is depicted in \Fig\ref{Fig:ANNmath}. It performs a linear transformation followed by a nonlinear activation:
\begin{equation}
\label{Eqn:ANN}
    V_j^i = f\left(\sum_{k=0}^{n_{i-1}} {W^i_{jk}} {V^{i-1}_k} + {b^i_j} \right),
\end{equation}
where $V_j^i$ denotes the activation of the $j^\text{th}$ neuron in the $i^\text{th}$ layer, ${W^i_{jk}}$ is the weight in the $i^\text{th}$ layer, applied from source neuron $k$ to destination neuron $j$ and $n_i$ represents the number of neurons in the $i^\text{th}$ layer. Hyperparameters, such as the number of neurons and the type of activation function, need to be defined to effectively extract relevant information from the input data.
%\subsubsection{Loss Function}

%The loss function used is the Mean Squared Error (MSE), defined as:
%\begin{equation}
%%\mathcal{L} = \frac{1}{N} \sum_{i=1}^{N} \left( y_i - \hat{y}_i \right)^2,
%\end{equation}
%where \( y_i \) is the predicted value, \( \hat{y}_i \) is the true value, and \( N \) is the number of training examples.

\subsection{Autoencoders}
An autoencoder is a neural network that learns a low-dimensional representation $\mathbf{z}\in\mathbb{R}^{r}$, termed the latent space, of a high-dimensional input $\mathbf{x}\in\mathbb{R}^{n}$ by aiming to reconstruct the input. It consists of an encoder that maps $\mathbf{x}$ to $\mathbf{z}$ $(\mathbf{z}= \phi(\mathbf{x}))$ and a decoder that maps $\mathbf{z}$ back to the high-dimensional space, $\mathbf{\tilde{x}}=\psi(\mathbf{z})$. The network is trained by minimising the reconstruction loss $\mathcal{L} = \left\|\mathbf{x} - \mathbf{\tilde{x}} \right\|^2$. In the current convolutional autoencoder (CAE) model, the encoder uses a 1D CNN to compress the Lamb-wave signal data into a latent space, and the decoder uses a transposed CNN (i.e., mirroring the encoder architecture) to reconstruct the input signal.

\subsection{Adam Optimization Algorithm}
The Adam optimisation algorithm~\citep{KingmaBa2014}, short for \textit{Adaptive Moment Estimation}, is a popular choice for optimising deep neural networks. It combines the benefits of two widely used optimisation methods: AdaGrad and RMSProp. Adam computes adaptive learning rates for each parameter by maintaining an exponentially decaying average of the gradients' first moment (mean) and the second moment (uncentered variance). 

At each time step \( t \), the Adam optimizer updates the parameters \( \theta_t \) based on the following equations:
\begin{equation}
g_t = \nabla_{\theta} J(\theta_t).
\end{equation}
The biased first- and second-moment estimates are updated according to
\begin{equation}
m_t = \beta_1 m_{t-1} + (1-\beta_1) g_t, \qquad
v_t = \beta_2 v_{t-1} + (1-\beta_2) g_t^2,
\end{equation}
where $\beta_1$ and $\beta_2$ denote the exponential decay rates for
the first and second moments, typically set to $\beta_1 = 0.9$ and
$\beta_2 = 0.999$, respectively. To eliminate initialization bias, the bias-corrected moment estimates are
given by
\begin{equation}
\hat{m}_t = \frac{m_t}{1-\beta_1^t}, \qquad
\hat{v}_t = \frac{v_t}{1-\beta_2^t}.
\end{equation}
Finally, the network parameters are updated as
\begin{equation}
\theta_{t+1} =
\theta_t -
\frac{\alpha\,\hat{m}_t}{\sqrt{\hat{v}_t} + \epsilon},
\end{equation}
where $\alpha$ is the learning rate and $\epsilon$ is a small constant
(typically $\epsilon = 10^{-8}$) introduced to ensure numerical stability.

\subsection{Feed Forward Neural Network}
\begin{figure}[!htp]
\centering
\includegraphics[trim = 1.5in 1.5in 1.5in 2.0in, clip=true, scale=0.50, angle=0,keepaspectratio]{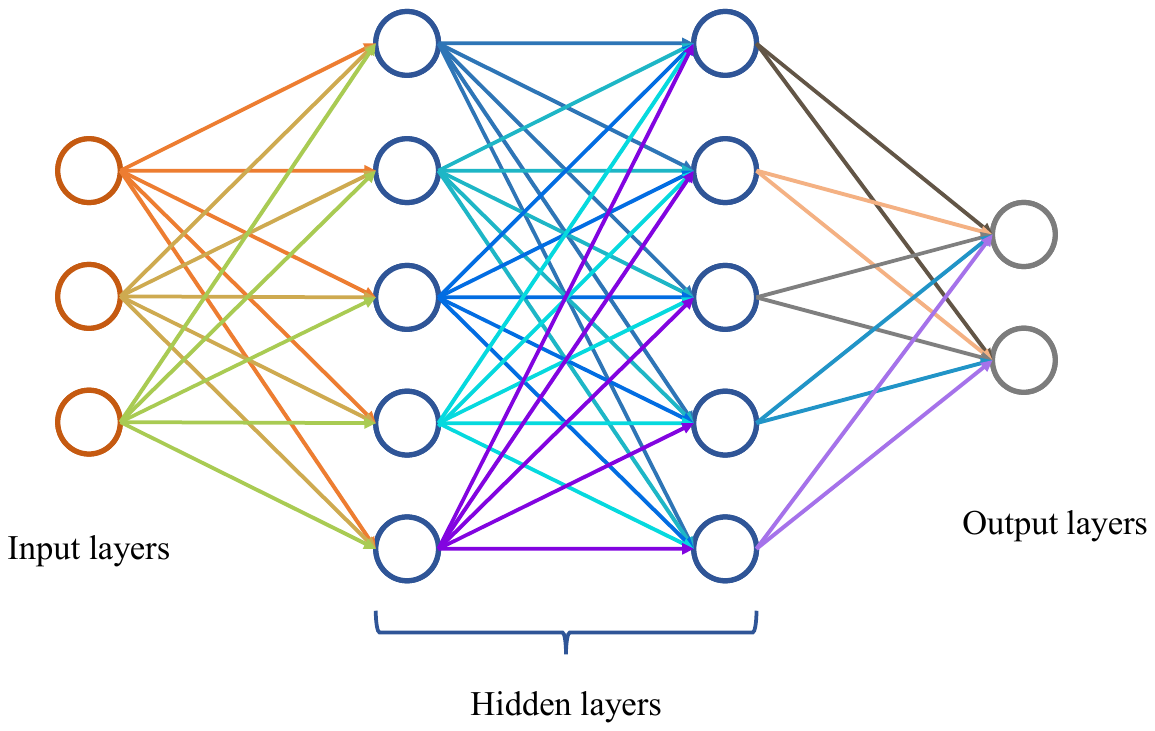}
\caption{Architecture of FFNN}\label{Fig:Architecture_FFNN}
\end{figure}

As illustrated in Fig.~\ref{fig:CAE-FFNN-TL-flowchart}, the latent representations of the Lamb wave signals are fed into an FFNN to predict the damage size and location. An FFNN consists of an input layer, one or more hidden layers, and an output layer, as shown in Fig.~\ref{Fig:Architecture_FFNN}. The information flows in one direction from the input to the output through these layers.

\begin{figure}[!htp]
    \centering
    \begin{minipage}{0.49\textwidth}
        \centering
        \includegraphics[trim = 0.7in 1.2in 1.1in 1.0in,width=\textwidth]{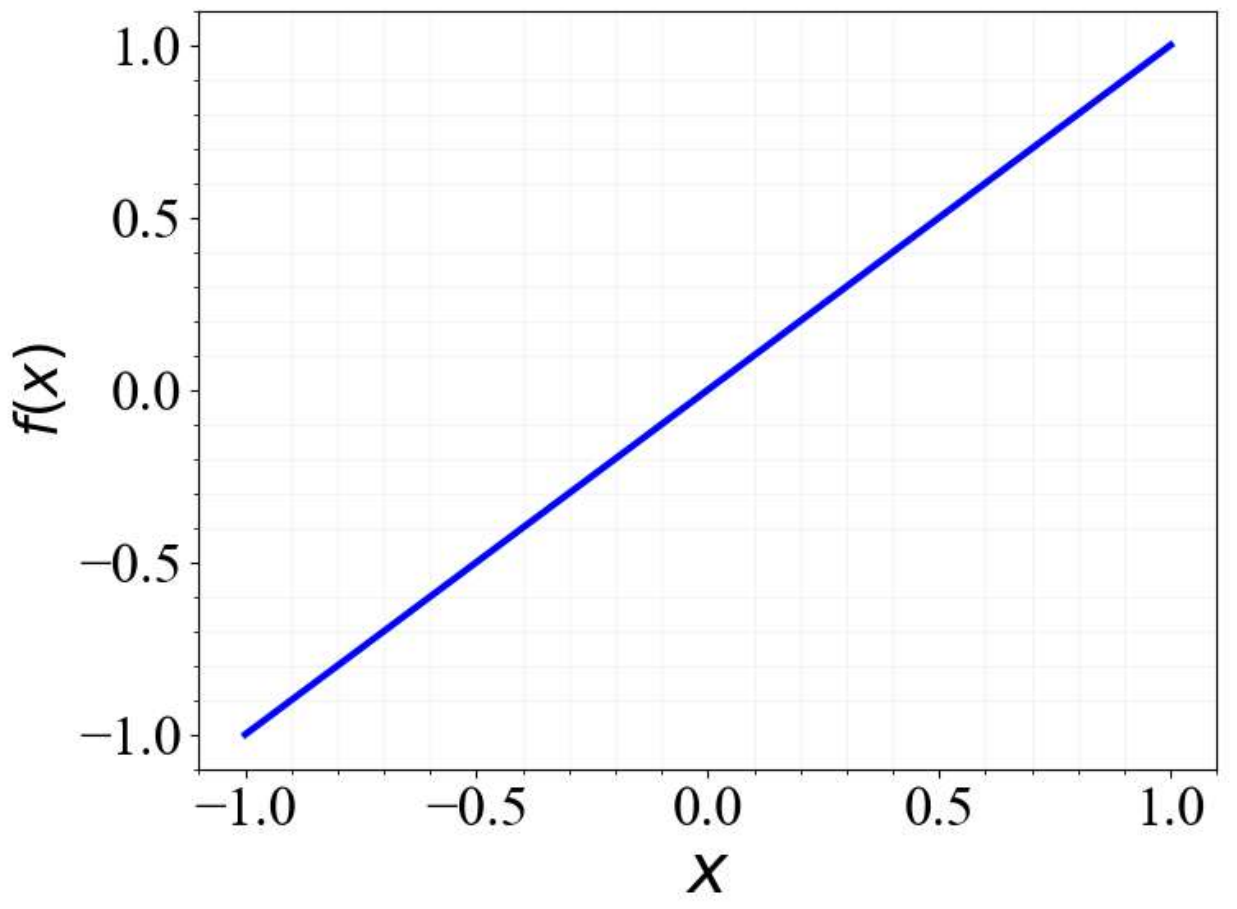}\\
        \small (a)
    \end{minipage}
    \hfill
    \begin{minipage}{0.49\textwidth}
        \centering
        \includegraphics[trim = 0.7in 1.2in 1.1in 1.0in,width=\textwidth]{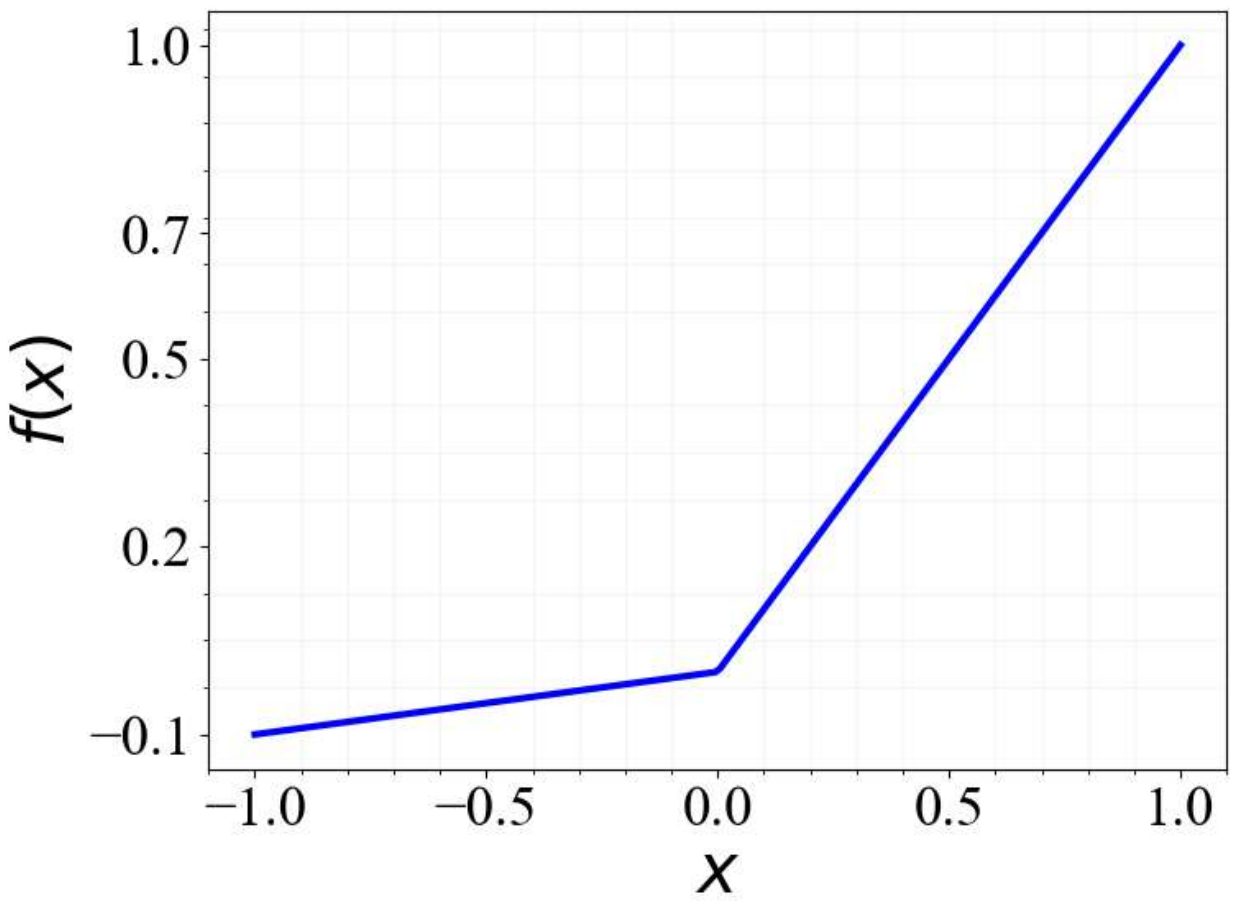}\\
        \small (b)
    \end{minipage}
    \caption{Activation functions: (a) Linear activation, (b) LeakyRelu activation.}
    \label{fig:Activation function}
\end{figure}

For each hidden layer $l$, the pre-activation output $\mathbf{z}_l$ is computed as
\begin{equation}
\mathbf{z}_l = \mathbf{W}_l\, f(\mathbf{x}_{l-1}) + \mathbf{b}_l,
\end{equation}
where $\mathbf{W}_l \in \mathbb{R}^{n_l \times n_{l-1}}$ is the weight matrix, $f(\mathbf{x}_{l-1})$ is the activation output of the previous layer (or the input \( \mathbf{x} \) if it is the first hidden layer), $\mathbf{b}_l \in \mathbb{R}^{n_l}$ is the bias vector. The Leaky ReLU activation function (\Fig\ref{fig:Activation function}) is applied to obtain the activation output \( \mathbf{a}_l=f({x}_l)\):
\begin{equation}
\mathbf{a}_l = \sigma_{\text{LeakyReLU}}(\mathbf{x}_l),
\end{equation}
where
\begin{equation}
\sigma_{\text{LeakyReLU}}(x) = \begin{cases} 
x, & \text{if } x > 0, \\
\alpha x, & \text{if } x \leq 0,
\end{cases},
\end{equation}
with $\alpha = 0.2$. The output layer employs the linear activation (\Fig\ref{fig:Activation function}). 

Having introduced the key components of the proposed CAE-driven transfer learning model, its architectural details for the current problem is presented next.

\subsection{CAE-driven transfer learning-based multi-fidelity model}
To enable accurate damage localisation under variable excitation frequencies within an SHM framework, a CAE-based transfer learning model is proposed. The model is initially trained on lightweight 1D-TDSE simulation data to learn robust, frequency-invariant feature representations and is subsequently fine-tuned using a limited set of high-fidelity experimental data via transfer learning.

\subsubsection{Base Model Training Using 1D TDSE-Simulated Lamb Wave Responses}
\begin{figure}[!ht]
\centering
\includegraphics[scale=0.65, trim = 0.3in 1.8in 0in 1.5in, clip=true, angle=0,keepaspectratio]{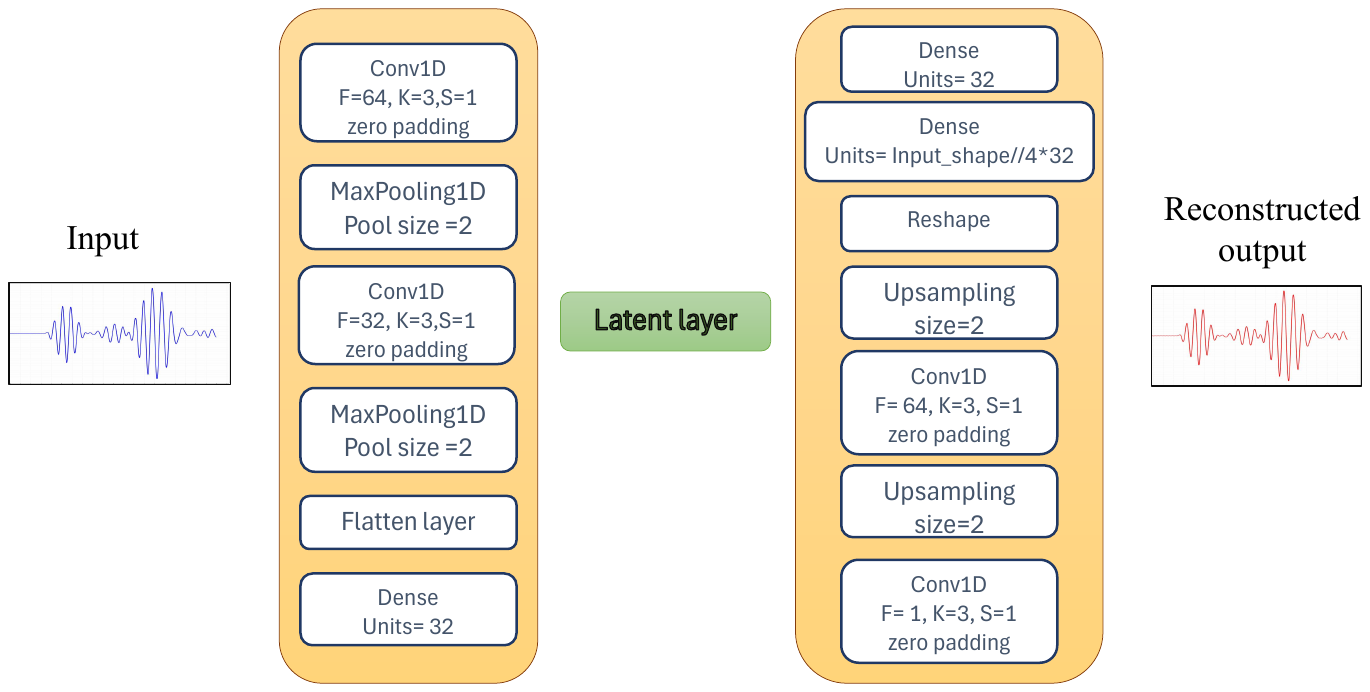}
\caption{Architecture of convolutional autoencoder}\label{Fig:Architecture_autoencoder}
\end{figure}

The CAE model is initially pretrained on a large dataset of Lamb wave forward-response signals generated from 1D-TDSE simulations, enabling robust feature learning from low-fidelity data. It is designed to learn a compact yet informative representation of the input signals while preserving their underlying physical characteristics. The architecture of the CAE is shown in \Fig\ref{Fig:Architecture_autoencoder}. The encoder consists of two 1D convolutional layers, each followed by a max pooling operation, to progressively reduce dimensionality and extract dominant patterns from the Lamb wave responses. The encoded features are flattened and passed through a dense layer to form a latent space, which captures the most relevant information for damage characterisation. 
The latent-space dimension, $n_L$, is chosen according to the variability of the input domain. For datasets corresponding to a single excitation frequency, $n_L$ is taken as 20, whereas for datasets containing multiple excitation frequencies, it is increased to 60 to capture the additional frequency-dependent characteristics of the signals. The layer-wise architectural details of the CAE framework for the variable-frequency dataset are provided in Table~\ref{tab:cae_architecture}. For the single-frequency dataset, the number of trainable parameters is 660 in the encoder latent layer and 672 in the first dense layer of the decoder. All other layer dimensions and trainable parameters remain unchanged.
\begin{table}[!ht]
\centering
\caption{Layer-wise architectural details of the convolutional autoencoder for variable-frequency dataset}
\label{tab:cae_architecture}

\begin{tabular}{lcc|lcc}
%\toprule
\hline
\multicolumn{3}{c|}{\textbf{Encoder}} &
\multicolumn{3}{c}{\textbf{Decoder}} \\
%\cmidrule(r){1-3} \cmidrule(l){4-6}
\hline

\textbf{Layer} & \textbf{Output shape} & \textbf{Params} &
\textbf{Layer} & \textbf{Output shape} & \textbf{Params} \\
\hline
Input Layer      & $(1000,1)$    & 0       &
Input Layer      & $(60)$        & 0 \\

Conv1D           & $(1000,64)$   & 256     &
Dense            & $(32)$        & 1,952 \\

MaxPooling1D     & $(500,64)$    & 0       &
Dense            & $(8000)$      & 264,000 \\

Conv1D           & $(500,32)$    & 6,176   &
Reshape          & $(250,32)$    & 0 \\

MaxPooling1D     & $(250,32)$    & 0       &
UpSampling1D     & $(500,32)$    & 0 \\

Flatten          & $(8000)$      & 0       &
Conv1D           & $(500,64)$    & 6,208 \\

Dense            & $(32)$        & 256,032 &
UpSampling1D     & $(1000,64)$   & 0 \\

Latent Layer     & $(60)$        & 1,980   &
Output Conv1D    & $(1000,1)$    & 193 \\

\hline

\multicolumn{2}{c}{\textbf{Total parameters}} &
\textbf{264,444} &
\multicolumn{2}{c}{\textbf{Total parameters}} &
\textbf{272,353} \\

\hline
\end{tabular}
\end{table}
The decoder mirrors the encoder architecture and reconstructs the input signal using two Conv1D layers combined with upsampling operations, ensuring that the latent representation retains the essential structural and propagation characteristics of the original Lamb wave signals. The CAE is trained using the Adam optimiser with a learning rate of 0.001 and a batch size of 8 for 450 epochs, balancing convergence and overfitting. Mean squared error is used as the loss function to minimise the reconstruction error between the input and reconstructed signals.

For downstream prediction, the encoder's latent representation is used as input to an FFNN that estimates the notch's location and depth. When the input signals are acquired for multiple central frequencies, the latent representation for each input signal is extracted from the latent space and concatenated with the corresponding excitation frequency. The combined data is used as input to the FFNN. The FFNN comprises an input layer of dimension $n_L$ and $n_L+1$ for the single- and variable-frequency cases, respectively, followed by fully connected dense layers, dropout layers, and a final output layer with two neurons representing the damage depth and location. The number of neurons and trainable parameters in each layer for the variable-frequency case are presented in Table~\ref{tab:ffnn_architecture}. This architecture enables effective modelling of the nonlinear relationship between the learned latent features and the corresponding damage characteristics.
\begin{table}[!htb]
\centering
\caption{Layer-wise architecture of FFNN used for damage size and damage location prediction from the encoded latent representation}
\label{tab:ffnn_architecture}

\begin{tabular}{lcc}
\hline
\textbf{Layer} & \textbf{Output shape} & \textbf{Parameters} \\
\hline
Input Layer & $(61)$ & 0 \\
Dense (ReLU) & $(128)$ & 7,936 \\
Dense (ReLU) & $(128)$ & 16,512 \\
Dropout & $(128)$ & 0 \\
Dense (ReLU) & $(32)$ & 4,128 \\
Output Layer & $(2)$ & 66 \\
\hline
\multicolumn{2}{r}{\textbf{Total trainable parameters}} &
\textbf{28,642} \\
\hline
\end{tabular}
\end{table}

The FFNN is trained with the same optimisation settings to ensure consistent, stable learning. During the testing phase, only the encoder component of the trained CAE is employed to extract latent representations from previously unseen 1D TDSE-generated signals. These latent features, together with the excitation frequency in the variable-frequency case, are then passed through the trained FFNN to estimate the corresponding damage characteristics.

\subsubsection{Transfer Learning with Experimentally Acquired Signals}
To adapt the pretrained model to experimental Lamb wave signals, the encoder of the previously trained CAE is kept frozen, while the decoder is fine-tuned using a limited experimental dataset, as illustrated in \Fig\ref{fig:CAE-FFNN-TL-flowchart}. This strategy enables the model to capture domain-specific variations present in experimental measurements while preserving the fundamental features learned from simulation data. In another setting, the experimental dataset may be replaced with high-fidelity simulation data obtained using the continuum-based 2D/3D FE model. In this work, we first test the transfer learning model on a limited dataset of 2D FE simulations from the \texttt{ABAQUS} software, before applying it to the experimental dataset. 

The decoder is retrained using the Adam optimiser with a reduced learning rate of 0.0001 to ensure stable convergence under limited data. Training is conducted for 500 epochs using the mean squared error loss function to minimise reconstruction errors between the experimental inputs and the decoded outputs.
After decoder fine-tuning, the frozen encoder is employed to extract latent representations from the experimental signals. These latent features, combined with the corresponding excitation frequencies, are provided as inputs to the pretrained FFNN, which is fine-tuned using the experimental dataset to predict notch depth and location. During the testing phase, unseen experimental signals are processed through the frozen encoder to obtain latent features, which, together with the excitation frequency, are subsequently passed to the fine-tuned FFNN for accurate estimation of the damage parameters.

\section{Results and Discussion}
In this section, the proposed CAE-FFNN-TL model is evaluated for predicting the size and location of notch-type damage in an aluminium plate from experimental measurements. Before training and testing the framework on experimental data, the performance of the TL framework is first assessed using a large dataset of lightweight 1D-TDSE simulations and a relatively smaller dataset of high-fidelity continuum-based 2D FE simulations in \texttt{ABAQUS}.

\subsection{Assessment with large 1D TDSE dataset and small 2D FE dataset}

\subsubsection{Data generation}
\label{sec: Data_generation}
To facilitate the development and training of the proposed multi-fidelity CAE-FFNN-TL model, two datasets were created: one derived from 1D-TDSE simulations and the other from 2D FE simulations conducted in \texttt{ABAQUS} for transfer learning-based fine-tuning. The datasets comprise time-history signals from piezoelectric sensors generated by Lamb wave propagation for different notch locations and depths.

For generating the 1D-TDSE dataset, an aluminium plate of length 1800 mm was considered, with a PZT-5A actuator located 600 mm from the left boundary. The actuator--sensor distance was varied between 300 mm, 400 mm, and 500 mm to represent different Lamb wave propagation paths. To capture a range of damage scenarios, notch locations were specified at 200 mm, 300 mm, and 400 mm, while notch depths spanned nine discrete values ranging from 0 mm to 2.7 mm (Table~\ref{tab: 1D TDSE 100kHz data}). This dataset served as the primary source for model pretraining, enabling the network to learn the fundamental characteristics of Lamb wave propagation under varying damage configurations.
\begin{table}[!htp]
\centering
\caption{Parameters with varying notch position and depth for simulation dataset generation}
\begin{tabular}{ccc}
\hline
\textbf{Parameter} & \textbf{Unit} & \textbf{Value} \\ \hline
Plate length ($a$)& mm & 1800  \\ \hline
Actuator distance from left boundary ($d_{LA}$)& mm & 600 \\ \hline
Actuator-sensor distance ($d_{AS}$)& mm & 300, 400, 500  \\ \hline
Notch position ($N_p$) & mm & 200, 300, 400  \\ \hline
Notch depth ($N_d$) for 1D TDSE & mm & 0, 0.3, 0.6, 0.9, 1.5, 1.8, 2.1, 2.5, 2.7  \\ \hline
Notch depth ($N_d$) for  2D FE & mm & 0, 0.3, 0.6, 0.9, 1.2, 1.5, 1.8  \\ \hline
\end{tabular}
\label{tab: 1D TDSE 100kHz data}
\end{table}

A high-fidelity dataset was generated via 2D FE simulations in \texttt{ABAQUS} to improve model generalisation. The essential parameters, such as the plate length and actuator-sensor position, were maintained the same as those in the 1D TDSE dataset to ensure consistency. Nonetheless, the depths recorded in this dataset were confined to six specific values, spanning from 0 mm to 1.8 mm (Table~\ref{tab: 1D TDSE 100kHz data}). This dataset was utilised for fine-tuning pretrained models, ensuring that insights gained from the 1D TDSE simulations could be successfully transferred and adapted to a high-fidelity setting. The utilisation of transfer learning effectively connected low-fidelity and high-fidelity simulations, enhancing the precision of notch localisation and minimising dependence on costly 2D FE simulations.

\subsubsection{Lamb wave response from 1D TDSE simulation}
\begin{figure}[!htb]
\centering
\includegraphics[scale=0.7, trim = 0in 0.2in 0in 0.0in, keepaspectratio]{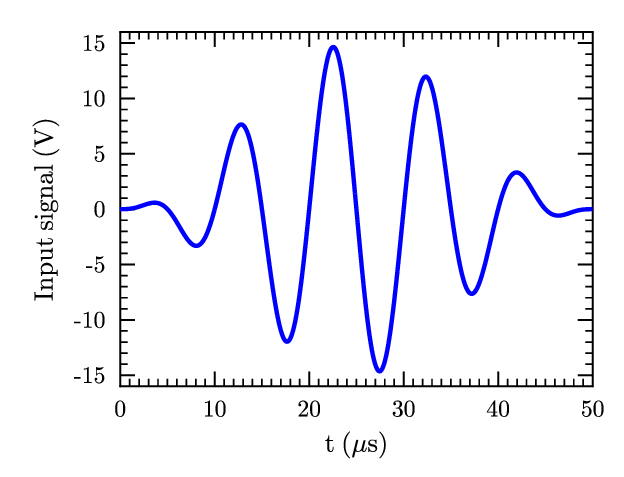}
\caption{Hann window-modulated five-cycle tone burst signal of 100 kHz central frequency}
\label{fig:TDSE_data_generation}
\end{figure}

The time history of the electric potential induced at the sensor's top surface is used to generate the dataset. The TDSE simulation is performed using the computer program developed in-house by \citet{JAIN2024}, as discussed in \Sec\ref{sec:1D TDSE Formulation}. The actuator is excited with a 5-cycle Hann window-modulated tone-burst signal of central frequency $f$, defined as
\begin{equation} 
\label{Eqn:Narrow band input}
V_A(t)=V_0/2(1-\cos{2\pi t/T_H})\sin{2\pi ft},
\end{equation}
where $V_0 $ denotes peak voltage, $T_H$ ($T_H = N_{B} /f$) represents Hann window length, and $N_{B}$ is the number of cycles in the tone burst. This input signal is commonly used in Lamb wave-based SHM applications to achieve a narrowband, well-localised wave packet that undergoes reduced dispersion during propagation, thereby facilitating easier identification of damage features. In this example, the central excitation frequency is taken to be $f=100$ kHz. The corresponding signal is shown in \Fig~\ref{fig:TDSE_data_generation}. The structure is discretised using 100 eight-noded ZIGT-based TDSE elements to obtain a converged solution for the sensor potential. The panel edges are subjected to free boundary conditions. The material properties considered for the model are presented in Table~\ref{Tab:matprop}. A time step of 0.1 $\mu$s is used for excitations with central frequencies up to 200 kHz, whereas a smaller time step of 0.05 $\mu$s is adopted for higher-frequency excitations (used later). This choice ensures at least 20 time steps per time period, providing sufficient temporal resolution to accurately capture Lamb wave propagation. 

\begin{table}[!htb]
\caption{Material properties used for simulations~\citep{KapuriaEtal2022,Sparkler}}
\begin{tabular}{ccccccccccc}
\hline
&        $Y_1$ &        $Y_2$ &        $Y_3$ &       $G_{23}$ &       $G_{13}$ &       $G_{12}$ &     $\nu_{12}$ &     $\nu_{13}$ &     $\nu_{23}$ &        $\rho$ \\
\cline{2-7} \cline{11-11}
  Material &                                                  \multicolumn{ 6}{c}{(GPa)} &            &            &            &   (kg/m$^3$) \\
\hline
 Al &       70.3 &       70.3 &       70.3 &      26.42 &      26.42 &      26.42 &       0.33 &       0.33 &       0.33 &       2700 \\

SP-5H &      66.67 &      66.67 &      47.62 &         23.0 &         23.0 &       23.5 &       0.29 &       0.51 &       0.51  &       7500 \\

% Gr/Ep$^3$ &      132.5 &      10.8 &      10.8 &     3.40 &         5.7 &         5.7 &       0.24 &       0.24 &       0.588  &       1600 \\

\hline
  &       $d_{31}$ &       $d_{32}$ &       $d_{33}$ &       $d_{15}$ &       $d_{24}$ &            &     $\eta_{11}$ &     $\eta_{22}$ &     $\eta_{33}$ &            \\
  
\cline{2-6} \cline{8-10}
           &                          \multicolumn{ 5}{c}{($\times$ $10^{-12}$ $mV^{-1}$)} &            &   \multicolumn{ 3}{c}{($\times$ $10^{-9}$ $F/m$)} &            \\

     SP-5H &       -265 &       -265 &        550 &        741 &        741 &            &      10.08 &      10.08 &      14.04 &            \\
\hline
\end{tabular}  
\label{Tab:matprop}

%$^1$~\cite{KapuriaEtal2022}, $^2$~\cite{Sparkler}
\end{table}

\subsubsection{Lamb wave response from 2D FE simulation}
For the high-fidelity continuum-based FE simulations, the plate is modelled in \texttt{ABAQUS} using eight-node biquadratic plane strain quadrilateral elements with reduced integration (CPE8R), while the piezoelectric transducers are modelled using eight-node biquadratic plane strain piezoelectric quadrilateral elements with reduced integration (CPE8RE). The mesh size for both the aluminium plate and the piezoelectric transducers is selected based on the commonly adopted criterion of at least 20 elements per shortest wavelength to ensure accurate numerical simulation of guided wave propagation~\citep{KapuriaEtal2022}.

\begin{figure}[!htp]
\centering
\includegraphics[scale=0.4, trim = 0.5in 2in 0in 1.5in,clip, keepaspectratio]{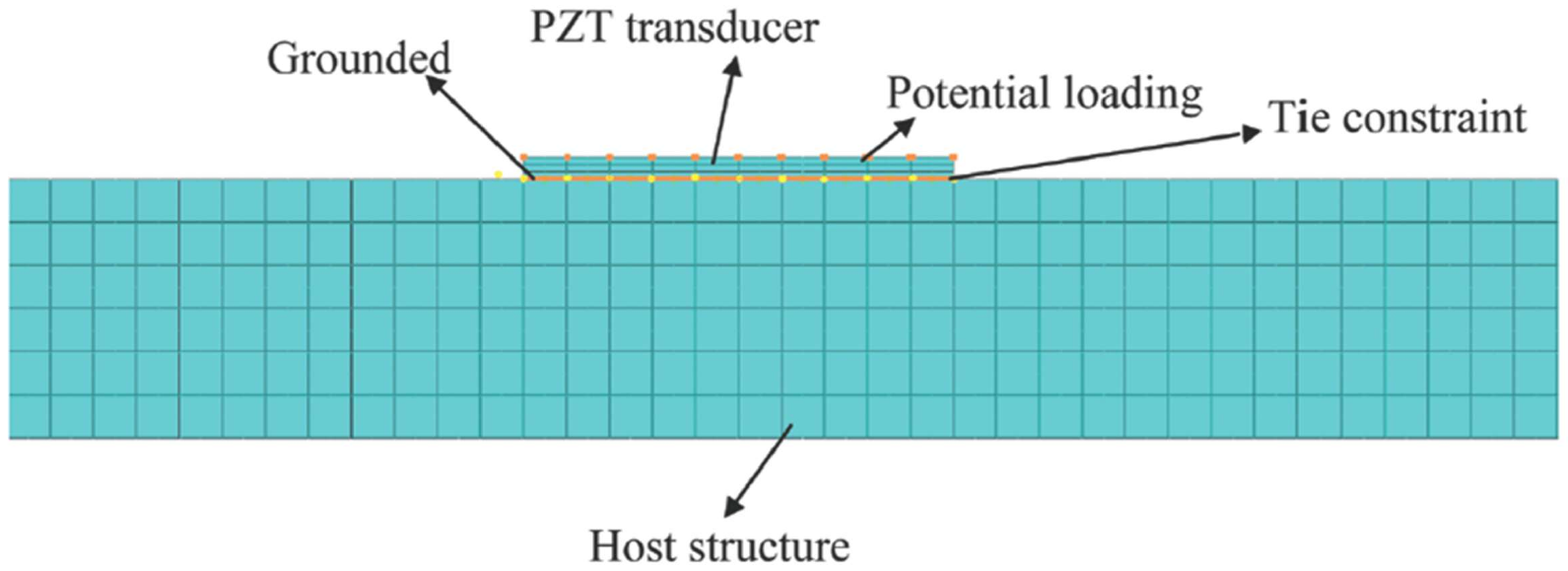}
\caption{Mesh configuration in 2D plane strain FE model}
\label{Fig:Mesh_config}
\end{figure}

In the thickness direction, the piezoelectric transducers and the plate are discretised using 2 and 8 elements, respectively. The element length in the longitudinal direction is 1 mm. The bottom nodes of the piezoelectric elements are connected to the top nodes of the elements in the host plate using the \texttt{TIE} constraint (\Fig~\ref{Fig:Mesh_config}). Electrically, the bottom surfaces of the piezoelectric transducers are grounded. An electric potential $V_A(t)$ is prescribed at the top surface of the actuator, while the voltage response is measured at the top electrode of the sensor by enforcing an equipotential constraint. Transient simulations are performed using the dynamic implicit integration scheme available in \texttt{ABAQUS} using the same time step as considered for the TDSE simulation. 

\subsubsection{Comparison of Lamb wave signals from 1D TDSE and 2D  FE simulations}
\begin{figure}[!htp]
    \centering
    \begin{minipage}{0.49\textwidth}
        \centering
        \includegraphics[trim = 0.4in 1.2in 0.5in 1.0in,width=\textwidth]{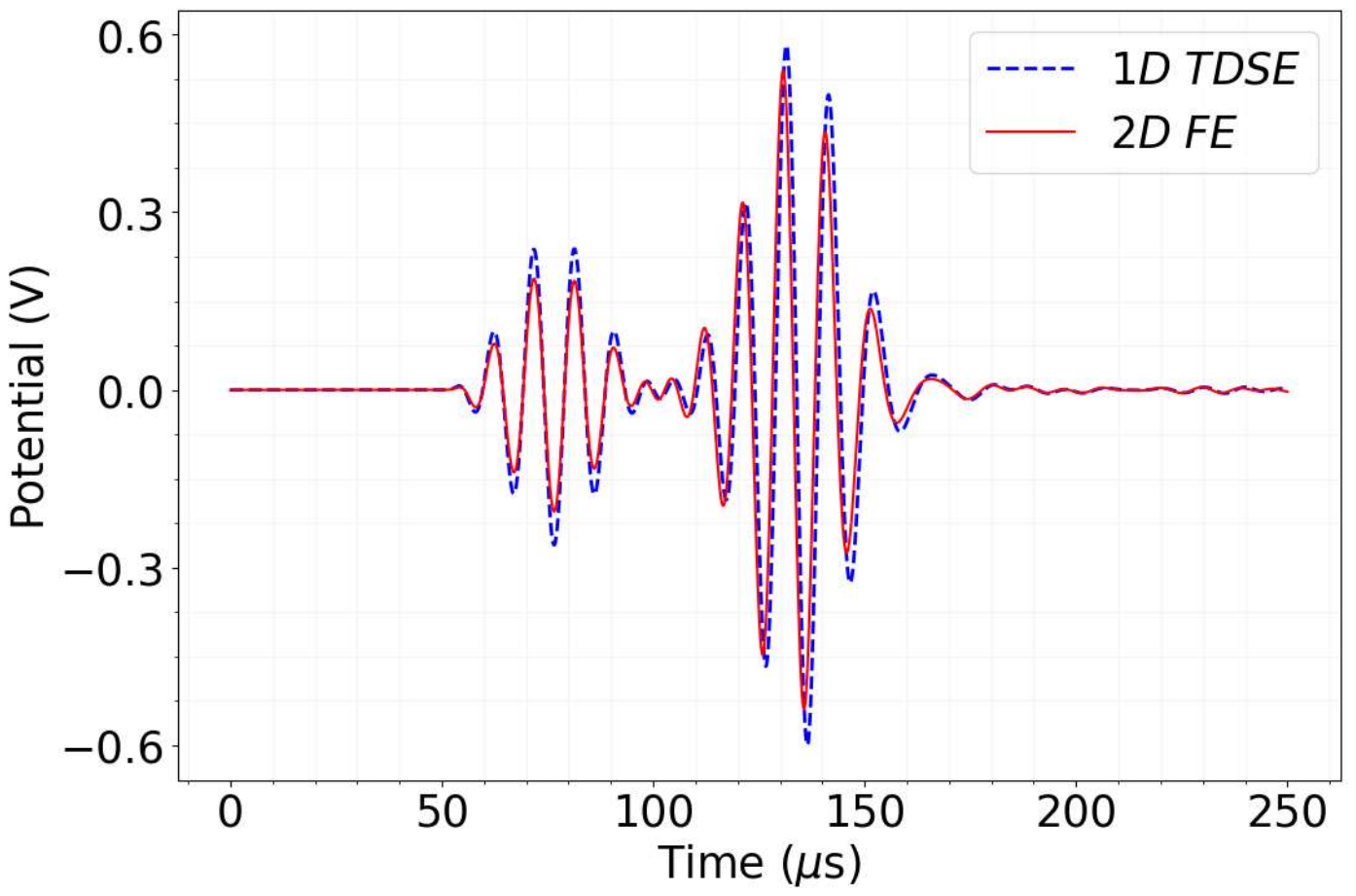}\\
        \small (a)
    \end{minipage}
    \hfill
    \begin{minipage}{0.49\textwidth}
        \centering
        \includegraphics[trim = 0.4in 1.2in 0.5in 1.0in,width=\textwidth]{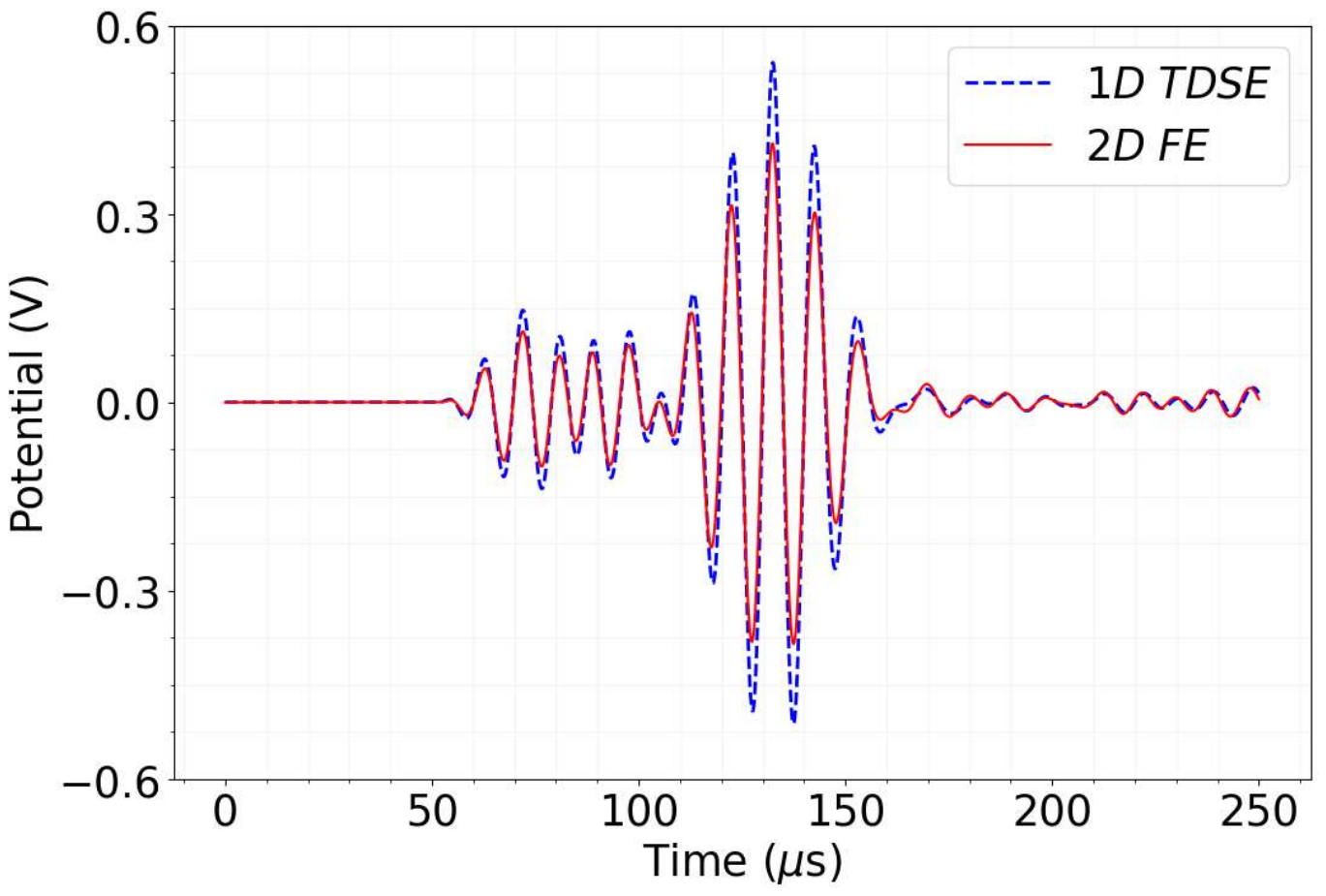}\\
        \small (b)
    \end{minipage}
    \caption{Comparison of 1D TDSE and 2D FE (\texttt{ABAQUS}) solutions for sensory potential for two values of notch depth: (a) 0.5 mm and (b) 2.0 mm.}
    \label{fig:Nd_comparison_ABAQUS}
\end{figure}

%\begin{figure}
%\centering
%\includegraphics[scale=0.55, trim = 0in 1.2in 0in 3.7in, keepaspectratio]{Al_Plate_config.pdf}
%\caption{Schematic diagram of the geometry of cross-section of aluminium plate}
%\label{Fig: plate_config}
%\end{figure}
Figure~\ref{fig:Nd_comparison_ABAQUS} illustrates the comparison between the time-history responses obtained from the two fidelity models for the aluminium plate-transducer configuration, where the centre-to-centre space between the actuator and the sensor ($d_{AS}$) is 280 mm, and the distance between notch and actuator ($d_{AN}$) is 70 mm. The sensor responses are shown for two notch depths, $ d_N = 0.5$ mm and 2 mm. A noticeable difference in the peak amplitudes of the fundamental symmetric mode $S_{0}$ and antisymmetric mode $A_{0}$ can be observed between the low-fidelity 1D TDSE and high-fidelity 2D FE results. The difference is higher for the larger notch depth. These results delineate the impact of model fidelity on the accuracy of wave mode responses and highlight the need for fine-tuning through transfer learning. 

\begin{figure}[!htb]
    \centering
    \begin{minipage}{0.49\textwidth}
        \centering
        \includegraphics[trim = 0.4in 1.2in 0.5in 1.0in,width=\textwidth]{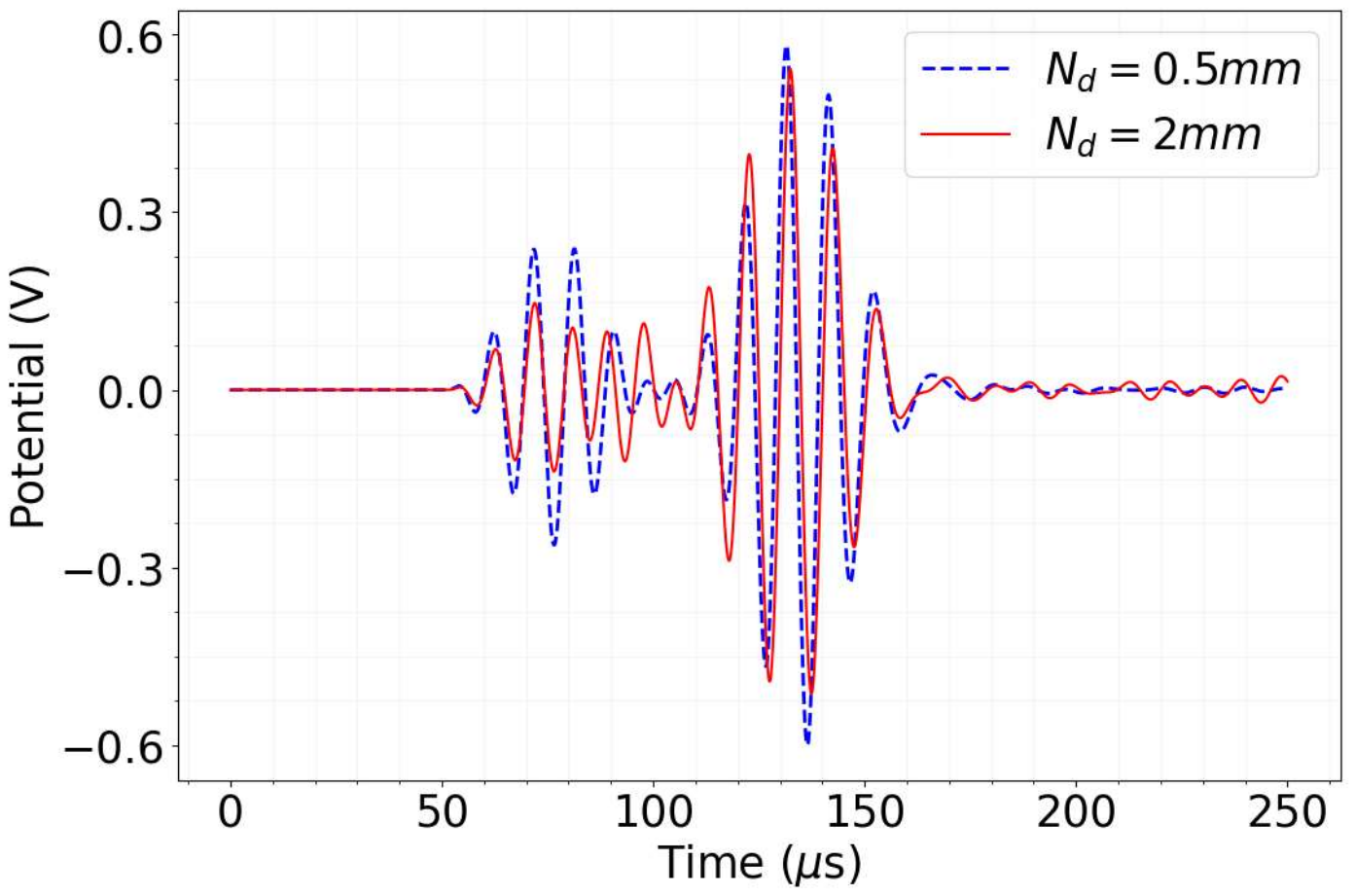}\\
        \small (a)
    \end{minipage}
    \hfill
    \begin{minipage}{0.49\textwidth}
        \centering
        \includegraphics[trim = 0.4in 1.2in 0.5in 1.0in,width=\textwidth]{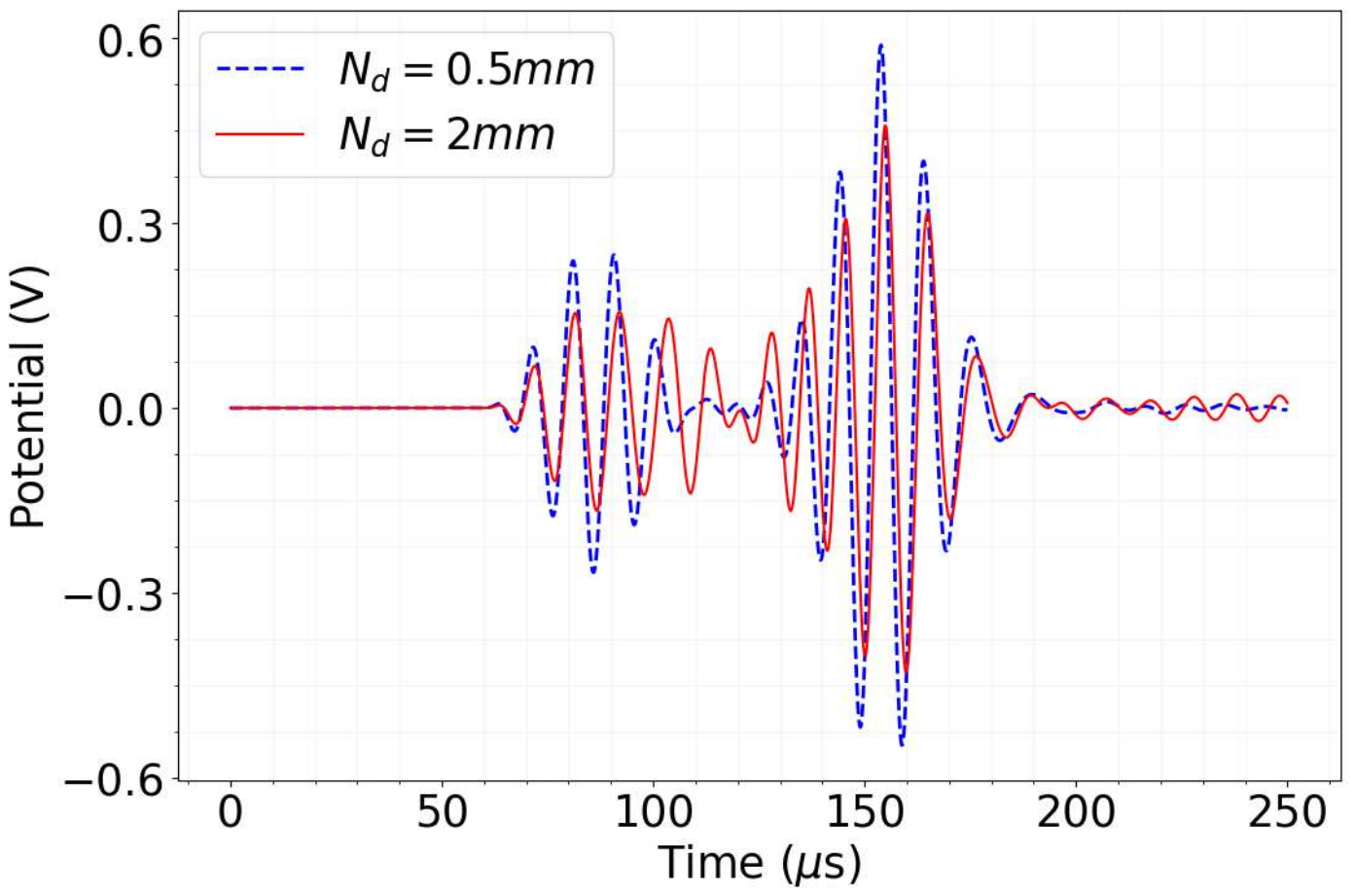}\\
        \small (b)
    \end{minipage}
    \caption{Effect of notch depth on Lamb wave-induced sensor potential (from 1D TDSE ) for notch positions of (a) $d_{AN}$= 70 mm and (b) $d_{AN}$= 85 mm.}
    \label{fig:notch_depth_comparison}
\end{figure}

The effect of notch depth on the Lamb wave signal is more clearly illustrated in \Fig\ref{fig:notch_depth_comparison} by plotting the sensor potentials obtained for two notch depths using 2D FE simulations. The comparison is made for two values of transducer spacings, $d_{AS}=280$ mm and 330 mm. The increase in the notch depth results in a reduction in the amplitudes of the $S_0$ and $A_0$ modes. The presence of the notch leads to the creation of an extra wave packet between them due to mode conversion resulting from interaction with the notch. 

\subsubsection{Evaluation of the proposed CAE-based transfer learning model}
The performance evaluation of the CAE-FFNN-TL model is conducted in two stages: initial training on 1D TDSE simulations and subsequent fine-tuning using 2D FE \texttt{ABAQUS} simulations. 
Table \ref{tab:dataset_split} presents the dataset split used for pretraining and fine-tuning phases of the CAE-driven TL model for notch detection.
This structured dataset split ensures that the model undergoes extensive training on 1D TDSE data before being fine-tuned with 2D FE \texttt{ABAQUS} data, enhancing their performance in damage localisation.
\begin{table}[!htp]
\setlength\tabcolsep{16pt}
    \centering
    \caption{Dataset split between training and testing phases for 1D TDSE and 2D FE simulation data}
    \renewcommand{\arraystretch}{1.2}
    \begin{tabular}{ccc}
        \hline
        & \multicolumn{2}{c}{No. of data samples}\\
        \cline{2-3}
        Phase & {Pretraining (1D TDSE)} & {Fine-tuning (2D FE-\texttt{ABAQUS})} \\
        \hline
        Training  & 307 & 152 \\
        \hline
        Testing  & 35 & 82 \\
        \hline
        Total  & 342 & 234 \\
        \hline
    \end{tabular}
    \label{tab:dataset_split}
\end{table}

The predictive performance of the DL models for notch localisation and depth is evaluated using the test samples by comparing the $ R^2$ score, defined as
\begin{equation}
R^2 = 1 - \frac{\sum_{i=1}^{n} (y_i - \hat{y}_i)^2}{\sum_{i=1}^{n} (y_i - \bar{y})^2},
\label{eq:r2_score}
\end{equation}
where $y_i$ represents the actual values, $\hat{y}_i$ represents the predicted values, $\bar{y}$ is the mean of actual values, $n$ is the total number of samples. An $R^2$ value close to 1 indicates a strong correlation between predicted and actual values, signifying high model accuracy, while an $R^2$ value close to 0 suggests poor predictive performance. The $R^2$ score of the CAE-FFNN model for predicting notch depth was 0.9972, indicating excellent accuracy, while the $R^2$ score for notch location prediction was 0.8880 (see Table~\ref{tab:cnnnet_vs_cae_tl}). These results confirm that the autoencoder effectively extracts relevant features, enabling the subsequent FFNN to achieve a higher accuracy in predicting notch characteristics.
\begin{table}[!htp]
\centering
\caption{Comparison of performances of present CAE-based TL model with CNN-based TL model}
\renewcommand{\arraystretch}{1.4}
\setlength{\tabcolsep}{10pt}
\begin{tabular}{ccccc}
\hline
{Model} & \multicolumn{2}{c}{{Pretraining $R^2$ score}} & \multicolumn{2}{c}{{Fine-tuning $R^2$ score}} \\
\cline{2-3} \cline{4-5}
 & {Notch depth} & {Notch location} & {Notch depth} & {Notch location} \\
\hline
{CAE-based} & 0.9972 & 0.8880 & 0.9941 & 0.9305 \\
\hline
{CNN-based} & 0.9758 & 0.3369 & 0.9871 & 0.8955 \\
\hline
\end{tabular}
\label{tab:cnnnet_vs_cae_tl}
\end{table}
  
\begin{figure}[!htb]
\centering
\includegraphics[scale=0.5, trim = 0in 1.2in 0in 1.2in, keepaspectratio]{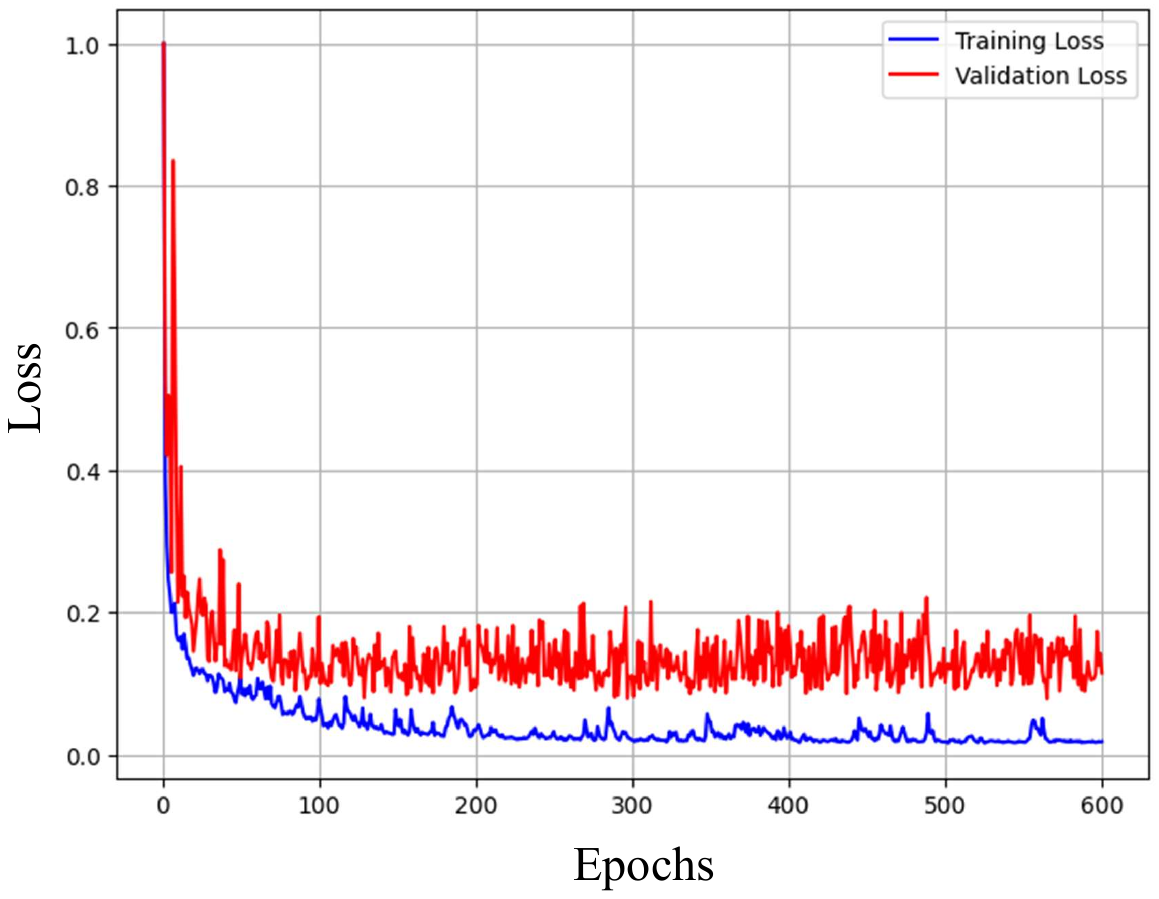}
\caption{Training performance graph for CAE-based TL model}
\label{fig:Loss_vs_Epochs}
\end{figure}

Transfer learning was employed to adapt the pretrained model to a high-fidelity data representation by fine-tuning it on a smaller dataset of 2D FE \texttt{ABAQUS} simulations, comprising 152 training and 82 test samples. Figure~\ref{fig:Loss_vs_Epochs} shows the variations of the loss over epochs for training and validation stages, confirming the adequacy of the selected number of epochs. The fine-tuned model achieved an $R^2$ score of 0.9941 for notch depth prediction, maintaining its high accuracy. More importantly, the $R^2$  score for notch location prediction improved to 0.9305, demonstrating that the model successfully adapted to high-fidelity simulation data. This significant increase in performance confirms the effectiveness of transfer learning in bridging the gap between low-fidelity synthetic data and high-fidelity numerical simulations, ensuring better localisation accuracy.

\begin{figure}[!htb]
\centering
\includegraphics[scale=0.5, trim = 0in 1.2in 0in 1.2in, keepaspectratio]{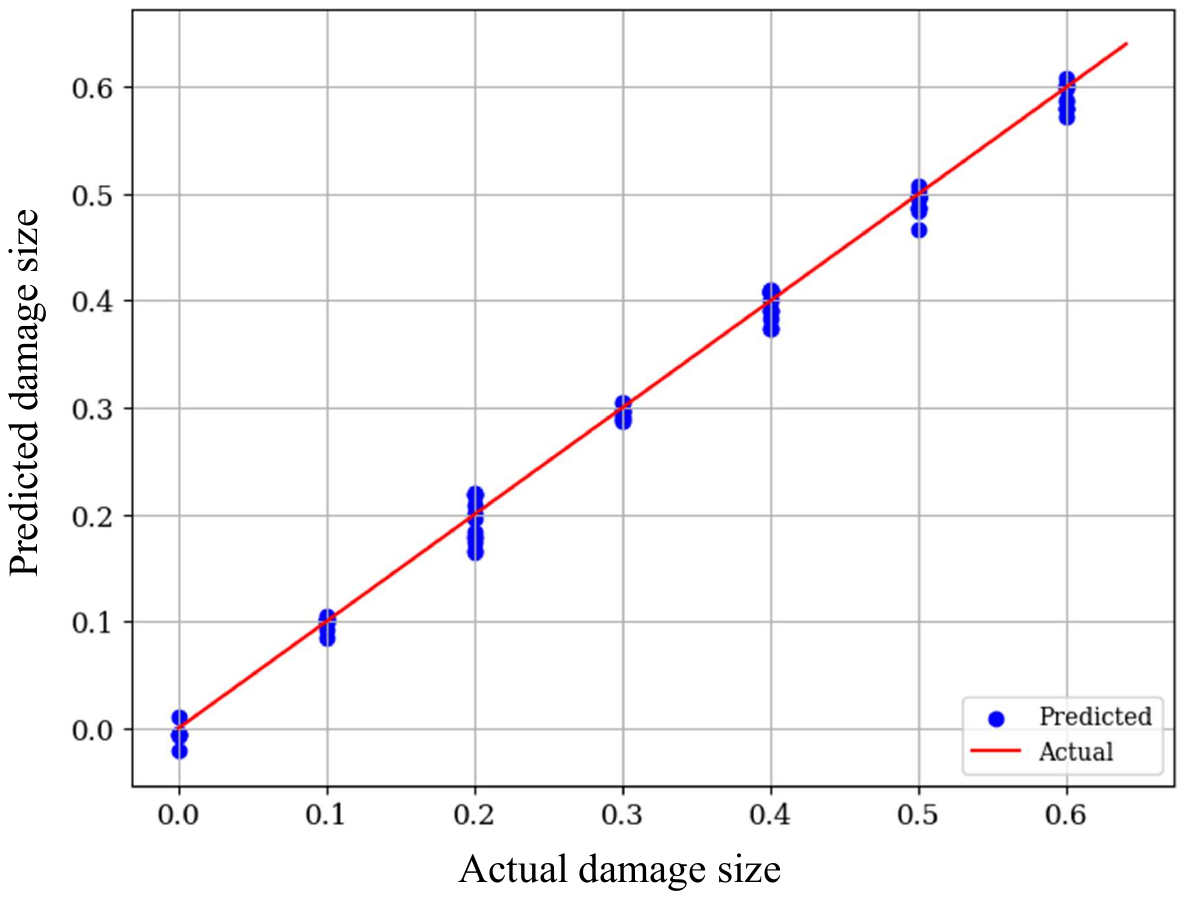}
\caption{Scatter plots of predicted and actual values of notch depth at the fine-tuning stage on 2D FE test data}
\label{fig: notch_depth_Autoencoder}
\end{figure}

The scatter plot in \Fig\ref{fig: notch_depth_Autoencoder} compares actual and predicted notch depths, evaluating the model's performance at estimating localised damage depth. The scatter points represent the model's predicted notch depths, closely aligning with the actual line. The evaluation results demonstrate the efficacy of the CAE-FFNN-TL model, revealing that combining deep feature extraction through autoencoding with predictive learning from FFNNs facilitates accurate notch localisation and depth estimation. Transfer learning enables the model to efficiently learn from a large synthetic dataset while effectively adapting to a high-fidelity dataset that represents practical damage characterisation scenarios.

\subsubsection{Comparison with CNN-based transfer learning model}
To illustrate the role of the latent space representation in the CAE framework, the proposed CAE-FFNN-TL model is compared with a deep learning model in which the CAE-based transfer learning architecture is replaced by a CNN-based transfer learning model, CNNnet-TL, proposed by Bao et al.~\citep{BaoEtal2023} for vibration-based condition monitoring. The CNNnet-TL architecture comprises four convolutional layers, each followed by a max pooling layer, to extract spatial features while progressively reducing dimensionality. These layers are followed by fully connected dense layers that refine the extracted features and facilitate damage characterisation. The final output layer predicts the notch depth and location based on the learned feature representations.

The model is implemented using the Keras library, optimised with the Adam optimiser, and trained with a learning rate of 0.001 for 600 epochs to ensure effective learning and convergence. Transfer learning is performed by freezing the convolution layers and fine-tuning the fully connected layers. The learning rate is set to 0.001, the batch size is 16, and the model is retrained for 600 epochs. The CNNnet-TL model is pretrained and fine-tuned using the same 1D TDSE-generated large dataset and the small 2D FE simulation dataset as the CAE-based TL model.

Table~\ref{tab:cnnnet_vs_cae_tl} shows a comparison of the performances of the present CAE-based TL model with the CNNnet-TL model through the $R^2$-score of predictions of notch depth and location. Although both models excelled at notch depth estimation, the CAE-based model showed markedly greater precision in notch localisation. The CNNnet-TL model produced an $R^2$-score of 0.3369 for notch localisation in the TDSE dataset, while the CAE-based model attained a significantly superior $R^2$-score of 0.8880. The performance disparity persisted after fine-tuning on the 2D FE data, with the CAE-based TL model achieving an enhanced localisation accuracy ($R^2$-score) of 0.9305, thereby affirming its capacity to extract and represent distinctive characteristics within the latent space.

\subsection{Assessment with large 1D TDSE dataset and small experimental dataset}
In practical SHM applications, the damage prediction must necessarily be performed on experimentally measured data from the actual structure. To illustrate this capability of the proposed model, the CAE-FFNN model pretrained on the large 1D TDSE dataset is now fine-tuned on a limited experimentally acquired Lamb wave response dataset. 

\subsubsection{1D TSDE data generation}
Unlike the previous example that considers only a single excitation frequency, this study considers three different excitation frequencies, 100 kHz, 150 kHz, and 200 kHz, for generating the dataset (Table~\ref{tab:Variable_freq_data}). All the geometrical details and boundary conditions are kept the same as described in \Sec 5.1.1 for the 100 kHz excitation case. But the mesh was made finer at higher frequencies, satisfying the spatial resolution requirement for converged results. 
\begin{table}[!htb]
\centering
\caption{Parameters with varying frequency, notch position, and notch depth for 1D TDSE for dataset generation}
\begin{tabular}{ccc}
\hline
\textbf{Parameter} & \textbf{Unit} & \textbf{Value} \\ \hline
Plate length ($a$)& mm & 1800  \\ \hline
Actuator distance from left boundary ($d_{LA}$)& mm & 600 \\ \hline
Actuator-sensor distance ($d_{AS}$)& mm & 300, 400, 500  \\ \hline
Excitation frequency ($f$) & kHz & 100, 150, 200  \\ \hline
Notch position ($N_p$) & mm & 100, 200, 250, 400  \\ \hline
Notch depth ($N_d$) & mm & 0, 0.3, 0.6, 0.9, 1.2, 1.5, 1.8, 2.1, 2.5, 2.7 \\ \hline
\end{tabular}
\label{tab:Variable_freq_data}
\end{table}

\begin{figure}[!htb]
    \centering
    \begin{minipage}{0.49\textwidth}
        \centering
        \includegraphics[trim = 0.7in 1.2in 0.6in 0.7in,width=\textwidth]{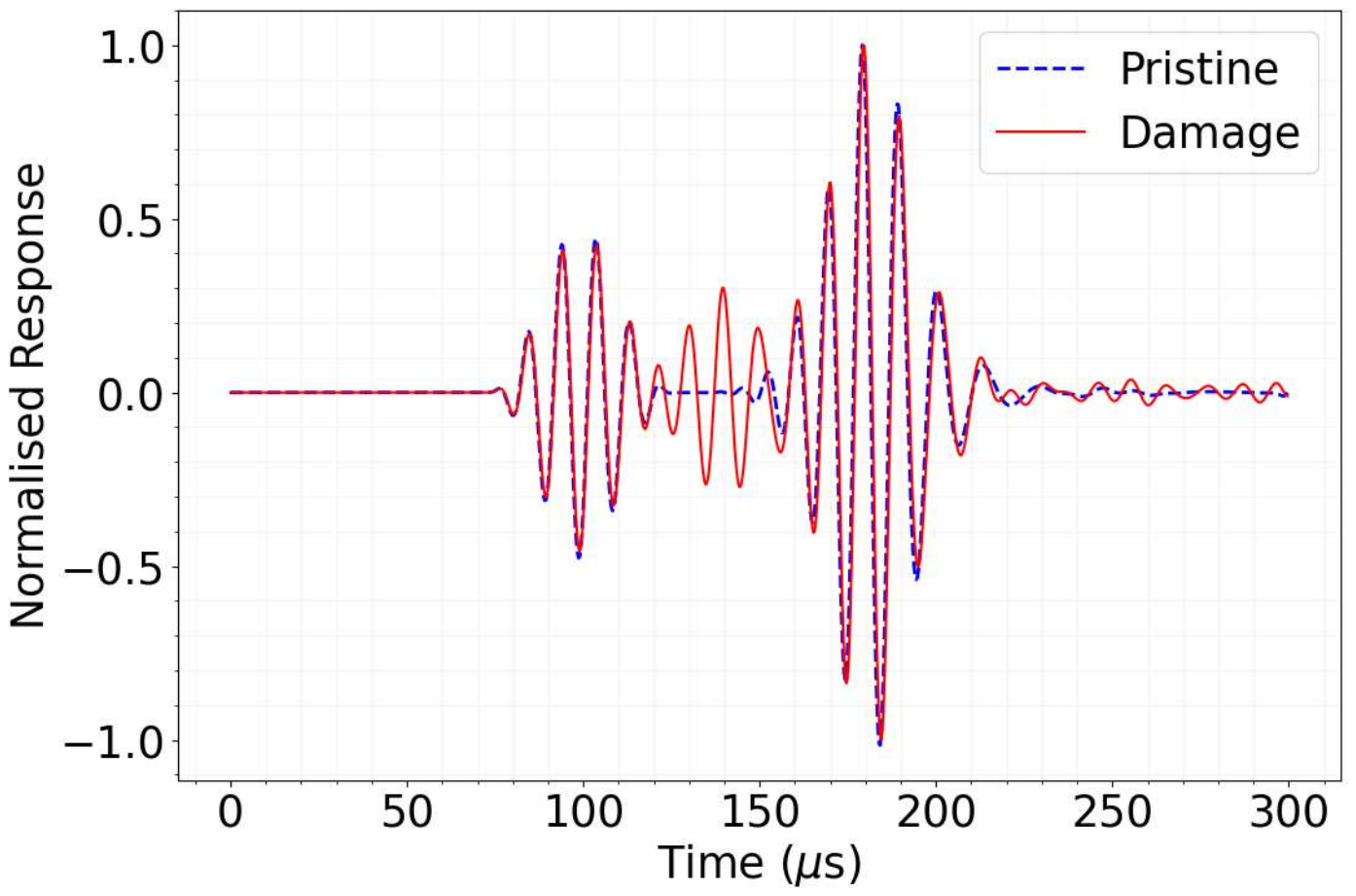}\\
        \small (a)
    \end{minipage}
    \hfill
    \begin{minipage}{0.49\textwidth}
        \centering
        \includegraphics[trim = 0.7in 1.2in 0.6in 0.7in,width=\textwidth]{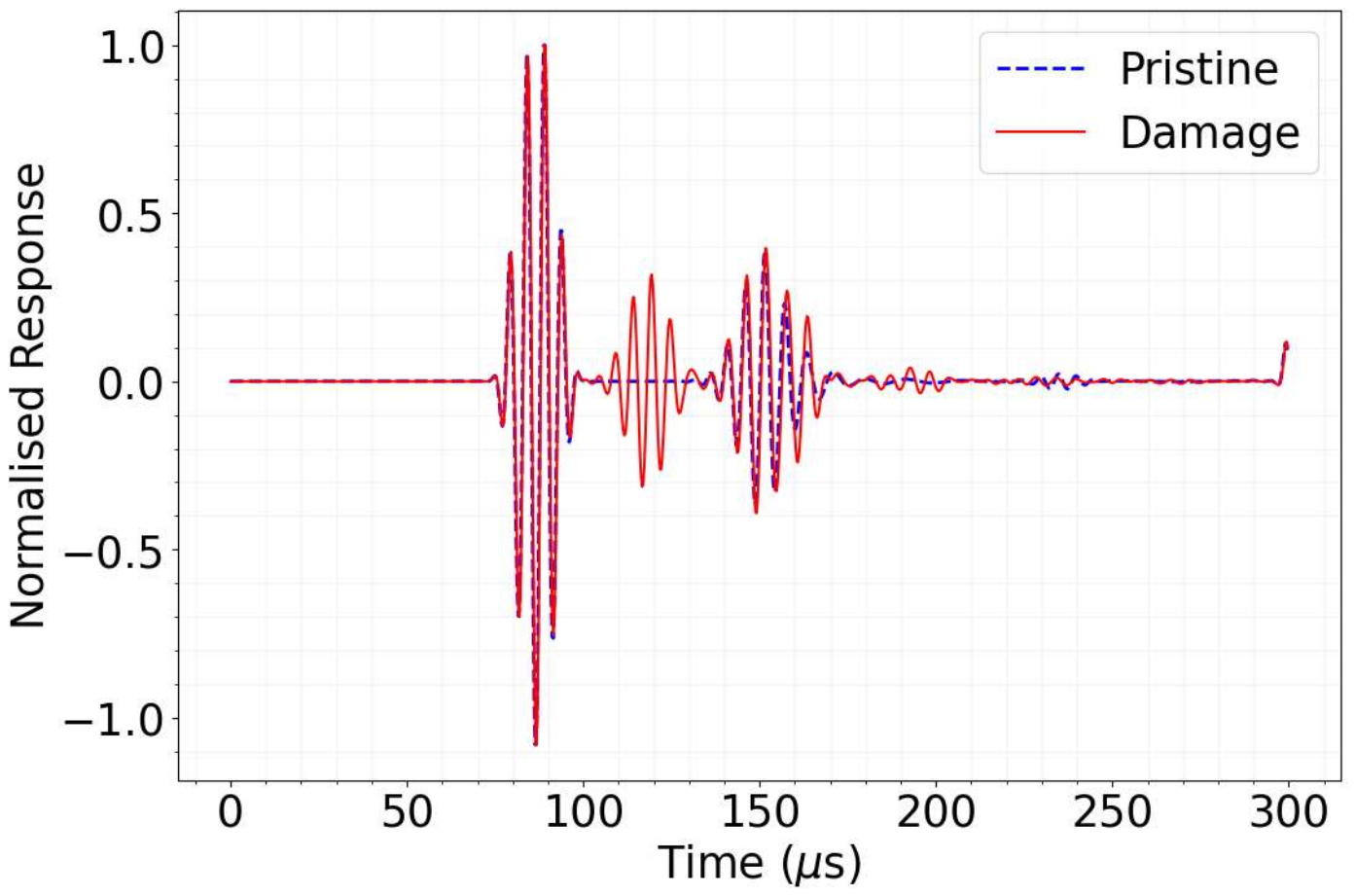}\\
        \small (b)
    \end{minipage}
    \caption{1D TDSE solution for sensor potential under (a) 100 kHz excitation and (b) 200 kHz excitation}
    \label{fig:100_&_200_kHz_freq}
\end{figure}

To illustrate the frequency dependence of response, a comparison of the time history of sensor potential for 100 kHz and 200 kHz excitation frequencies is presented in \Fig\ref{fig:100_&_200_kHz_freq}. It is evident that a higher frequency (200 kHz) has resulted in greater sensitivity to damage than a lower frequency (100 kHz). These results also clearly demonstrate how excitation frequency influences the amplitudes and arrival times of the direct Lamb wave modes ($S_0$ and $A_0$), as well as the damage-induced extra mode, highlighting the importance of incorporating multi-frequency data in ML model development.

For each excitation frequency, 66 Lamb wave forward responses were simulated using the 1D TDSE model, resulting in $66 \times 3 = 198$ base signals. To enhance the variability of the training data and improve the model's robustness to noise, each base signal was augmented with six levels of additive Gaussian noise, with signal-to-noise ratios (SNRs) ranging from 20 dB to 30 dB. This augmentation expanded the dataset by a factor of six, yielding a total of $198 \times 6 = 1188$ samples. Out of the complete TDSE dataset, 1069 samples were allocated for pretraining the CAE-FFNN model, while 119 samples were reserved for testing, as summarised in Table~\ref{tab:Exp_CAE_data_split}. 
\begin{table}[!htp]
    \centering
    \caption{Dataset split between training and testing phases for 1D TDSE simulation and experimental data}
    \renewcommand{\arraystretch}{1.2}
    \setlength{\tabcolsep}{15pt}
    \begin{tabular}{ccc}
     \hline
        & No. of data samples \\
        \cline{2-3}
        Phase & Pretraining (1D TDSE data) & Fine-tuning (Experimental data)\\
        \hline
        Training & 1069 & 168 \\
        \hline
        Testing & 119 & 24 \\
        \hline
        Total & 1188 & 192 \\
        \hline
    \end{tabular}
    \label{tab:Exp_CAE_data_split}
\end{table}

\subsubsection{Experimental data generation}
\begin{figure}[!htb]
\centering
\includegraphics[width=0.95\textwidth, 
                   trim=1cm 6cm 1cm 6cm, clip]{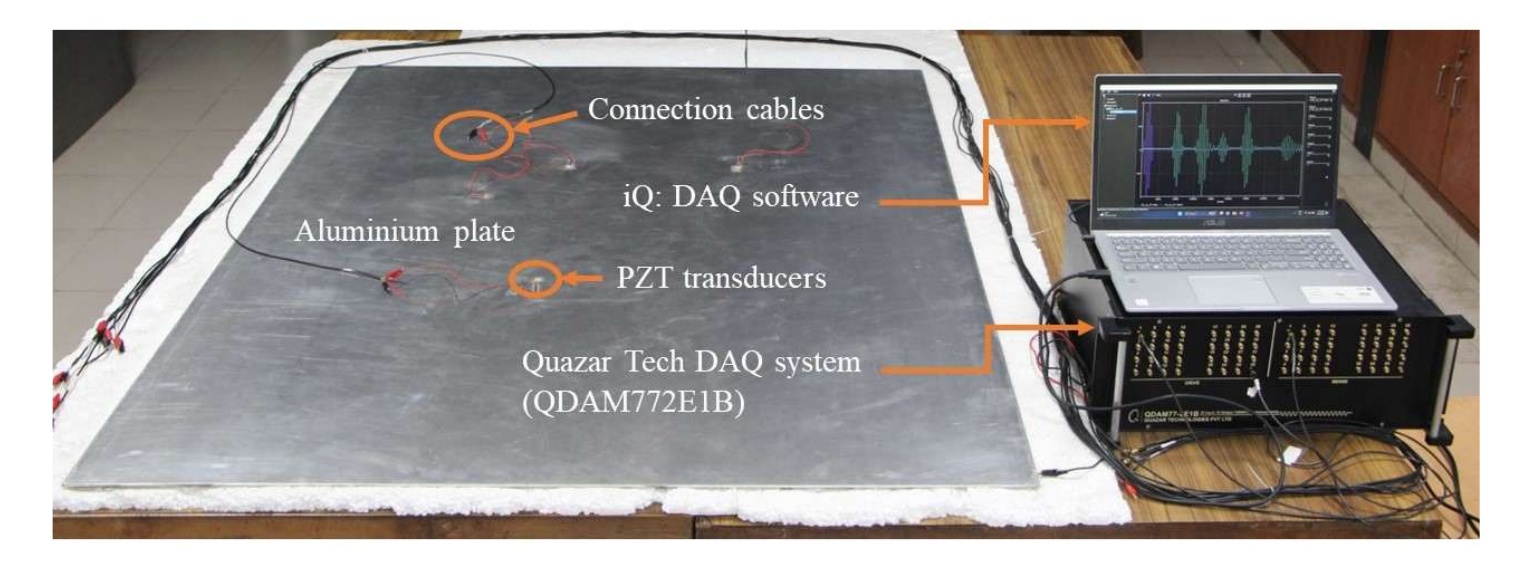}
\caption{Experimental setup illustrating the arrangement of PZT transducers and acquisition system for Lamb wave generation and sensing.}
\label{fig: Exp_config}
\end{figure}

The experiments were conducted on an aluminium plate with dimensions of $2438\,\text{mm} \times 1220\,\text{mm} \times 3\,\text{mm}$. The setup is shown in Fig.~\ref{fig: Exp_config}. Three wafer-type SP-5H PZT transducers, each of size $10\,\text{mm} \times 10\,\text{mm} \times 0.25\,\text{mm}$, were bonded to the surface of the plate using Araldite, a two-part epoxy adhesive. These PZTs were sourced from Sparkler Ceramics India. A combined function generator and data acquisition system (Model: QDAM772E1B) from Quazer Tech Pvt. Ltd., India, was used for both signal excitation and measurement. The system includes high-speed analogue-to-digital (A--D) and digital-to-analogue (D--A) conversion channels with 14-bit resolution, selectable gain settings, and programmable sampling frequencies. This configuration enables accurate generation, amplification, and recording of high-frequency Lamb wave signals in both pristine and damaged conditions.

\begin{figure}[!htb]
\centering
\includegraphics[scale=0.58, trim = 0in 1.2in 0in 1.2in, keepaspectratio]{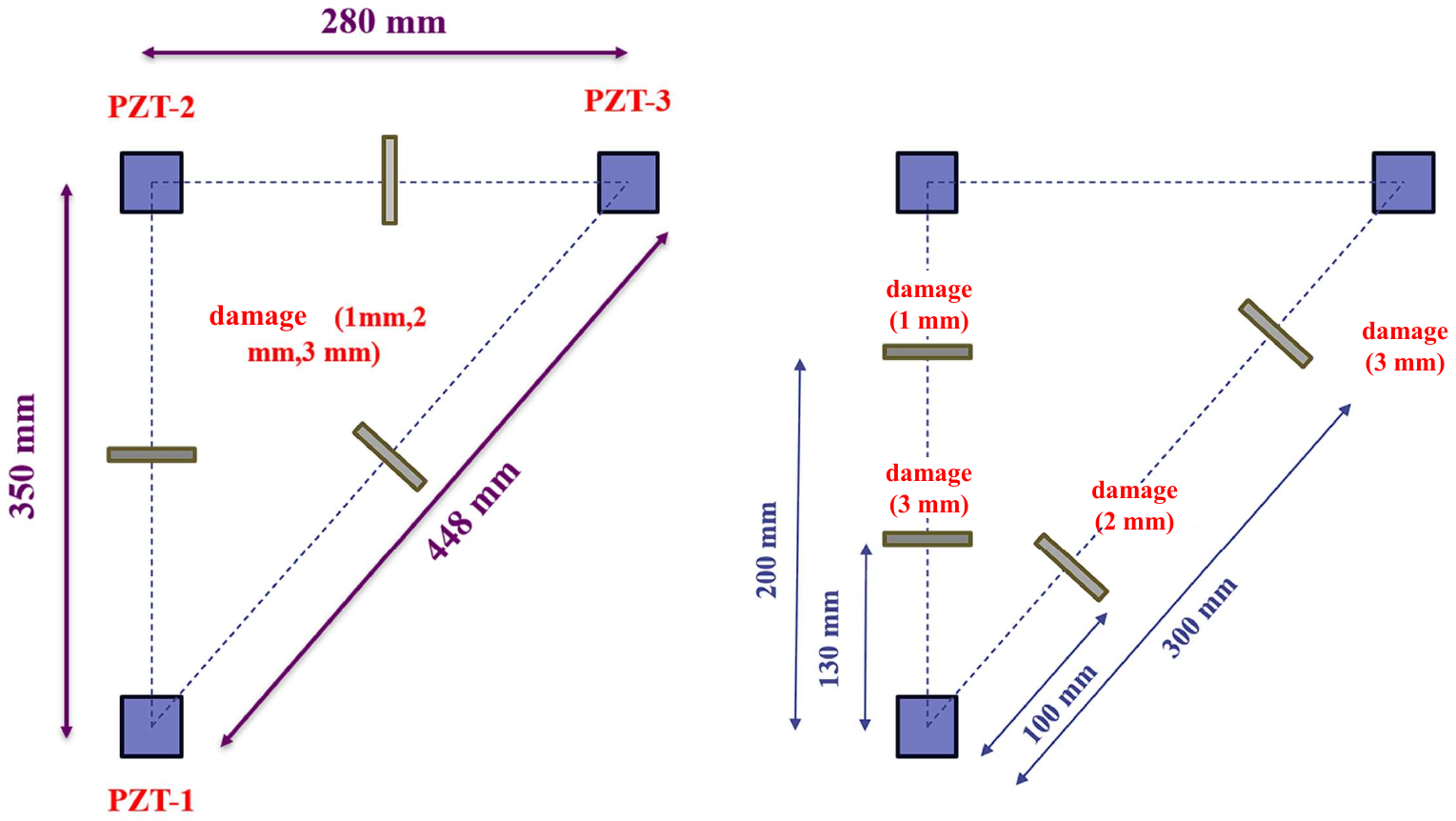}
\caption{Configuration of the PZT transducers and block-mass damage positions for generating experimental data}
\label{fig: PZT_config}
\end{figure}

As shown in \Fig\ref{fig: PZT_config}, the PZT transducers were bonded to the plate surface in a triangular configuration, with inter-sensor distances of 350 mm (PZT-1--PZT-2), 280 mm (PZT-2--PZT-3), and 448 mm (PZT-1--PZT-3). Damage scenarios were introduced in the form of aluminium strip block masses positioned along the wave propagation paths to evaluate the sensitivity of Lamb wave responses to defects. The strip heights were varied as 1, 2, and 3 mm to simulate different damage severities. A mid-path damage was introduced between PZT-1 and PZT-3 for all three damage sizes. Additionally, along the PZT-1--PZT-2 path, block masses with heights of 1 mm and 2 mm were placed at distances of 130 mm and 200 mm from the actuator, respectively. Similarly, along the PZT-1--PZT-3 path, block masses with heights of 2 mm and 3 mm were located at distances of 100 mm and 300 mm from the actuator, respectively. This configuration enabled the acquisition of a relatively limited dataset capturing the effects of damage size and location on  Lamb wave response signals.

\begin{figure}[!htp]
    \centering
    \begin{minipage}{0.49\textwidth}
        \centering
        \includegraphics[trim = 0.4in 1.2in 0.5in 1.0in,width=\textwidth]{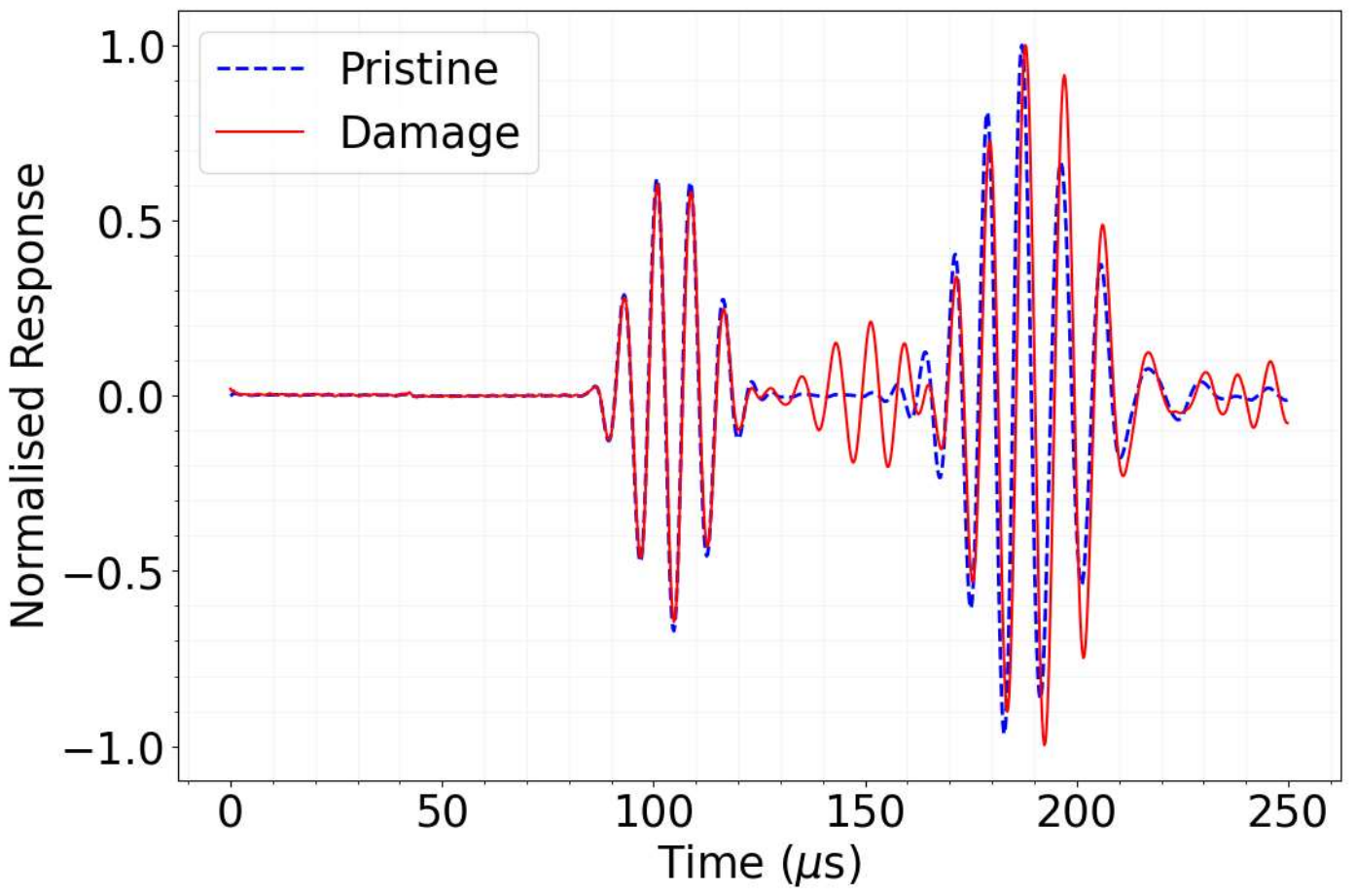}\\
        \small (a)
    \end{minipage}
    \hfill
    \begin{minipage}{0.49\textwidth}
        \centering
        \includegraphics[trim = 0.4in 1.2in 0.5in 1.0in,width=\textwidth]{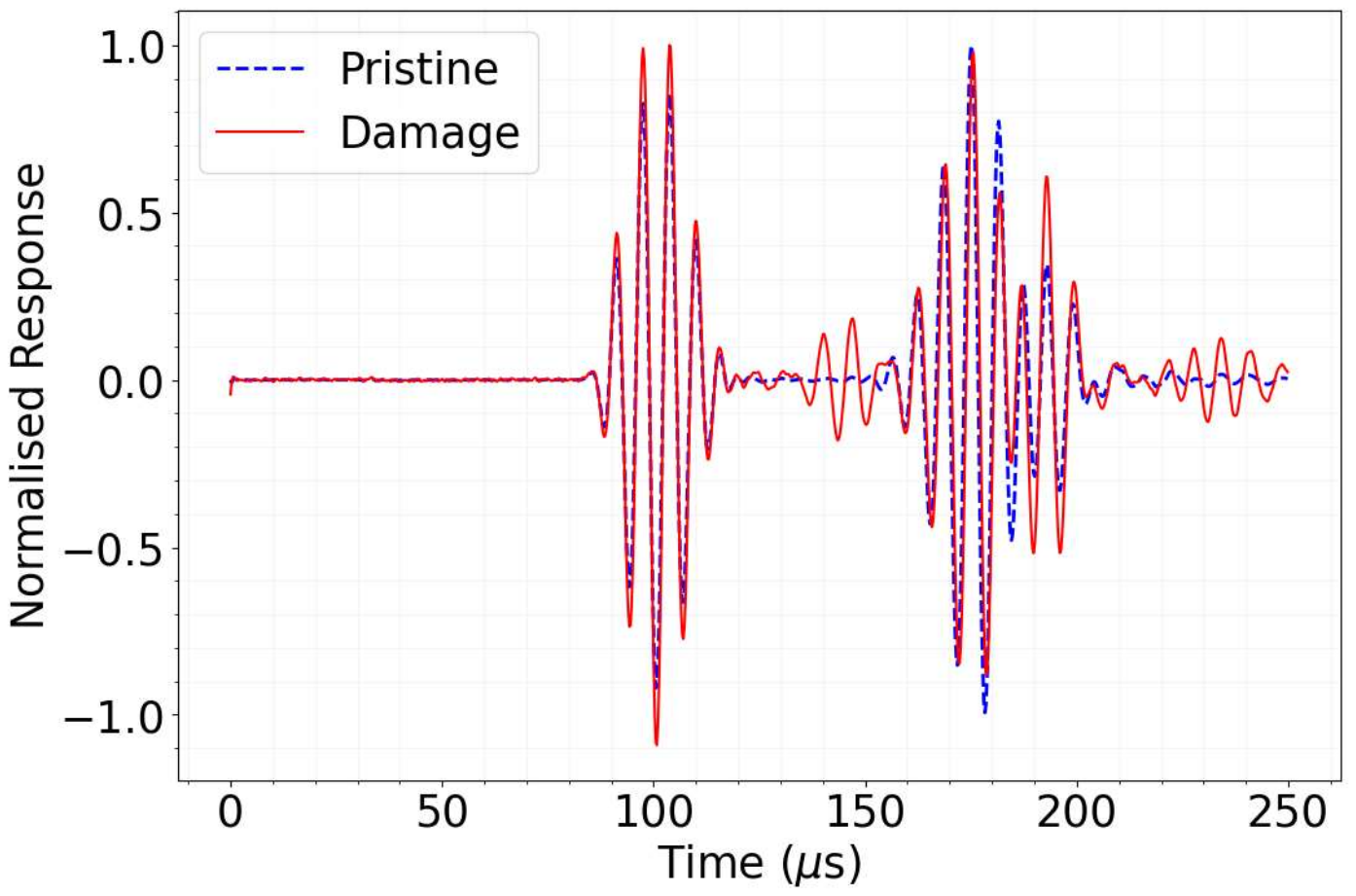}\\
        \small (b)
    \end{minipage}
    \caption{Experimental sensor signals for path PZT-1--PZT-3 and damage height of 2 mm under (a) 120 kHz and (b) 150 kHz excitations.}
    \label{fig: Exp_response}
\end{figure}

Lamb wave responses were generated by exciting one of the surface-bonded transducers using a Hann-window modulated tone-burst signal comprising five cycles with a peak voltage of 15 V. Signal acquisition was performed using an integrated data acquisition system configured with a gain of 20 dB and a sampling rate of 72 MS/s. To enhance the signal-to-noise ratio (SNR), each measurement was repeated 50 times and ensemble-averaged. The measured signals for the 445 mm-long sensing path, with and without damage, are shown in \Fig\ref{fig: Exp_response}, illustrating the dependence of the wave response and damage-induced features on the excitation frequency.

The experimentally acquired Lamb wave responses were used to fine-tune the pretrained model through transfer learning. As indicated in Table~\ref{tab:Exp_CAE_data_split}, a total of 192 experimental signals were collected under varying conditions of excitation frequency, notch depth, and notch location. From this dataset, 168 signals were used for fine-tuning, and the remaining 24 were used to test and evaluate the model's generalisation capability on real-world data.

\subsubsection{Evaluation of the CAE-FFNN model trained on large simulated data}
\begin{figure}[!htp]
    \centering
    \begin{minipage}{0.49\textwidth}
        \centering
        \includegraphics[trim = 1.1in 1.2in 1.0in 1in,width=\textwidth]{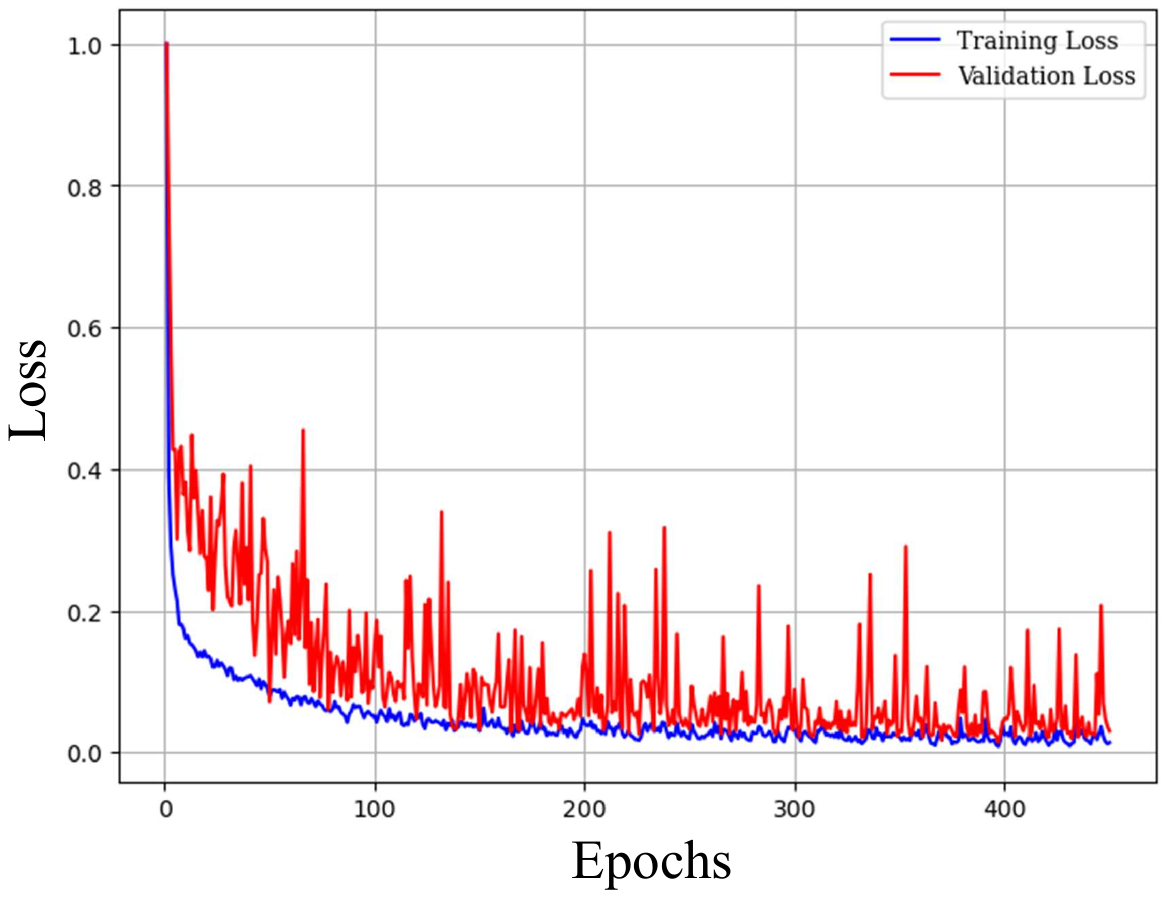}\\
        \small (a)
    \end{minipage}
    \hfill
    \begin{minipage}{0.49\textwidth}
        \centering
        \includegraphics[trim = 1.0in 1.2in 1.1in 1in,width=\textwidth]{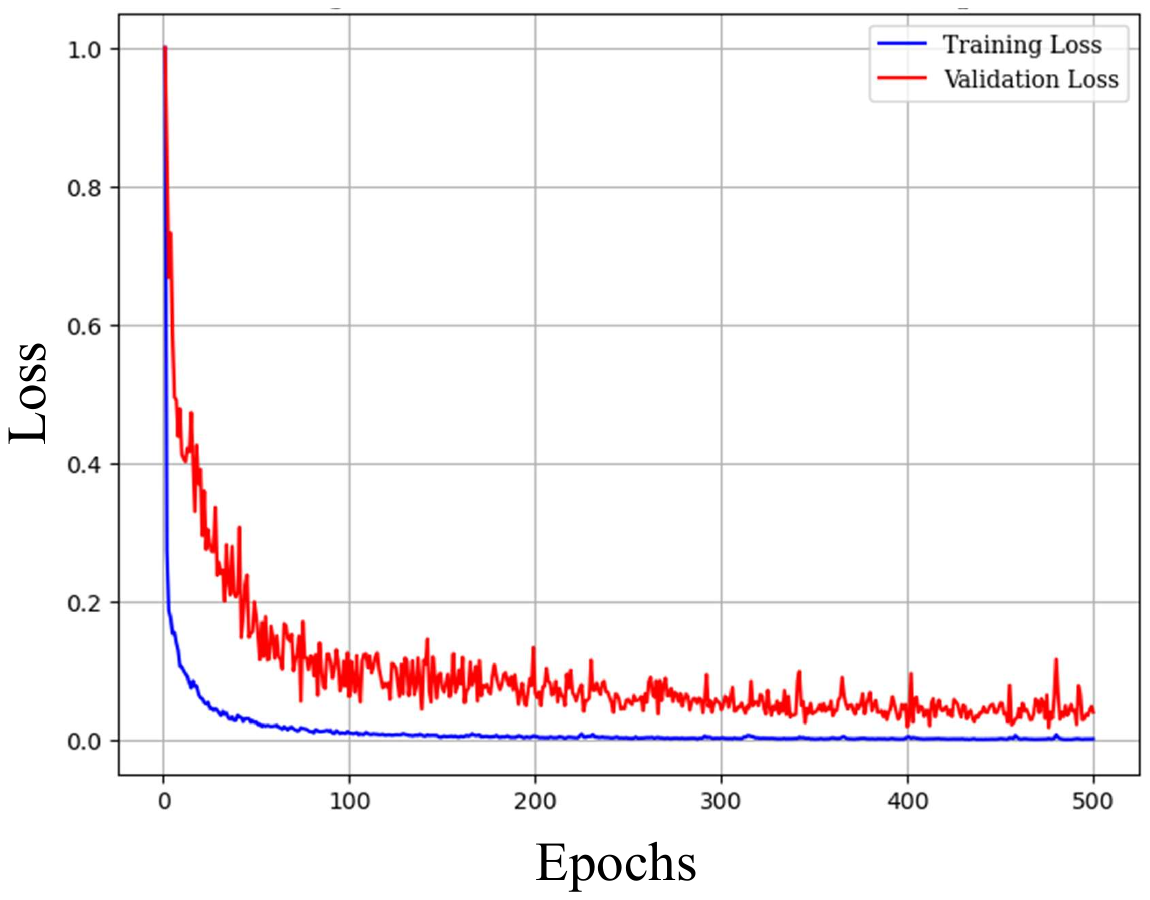}\\
        \small (b)
    \end{minipage}
\caption{Training performance of CAE-FFNN-TL model for damage localisation and sizing at (a) pretraining stage with a large 1D TDSE simulation dataset and (b) fine-tuning stage with few experimental data}
    \label{fig:Loss_vs_Epochs_case_2}
\end{figure}

The performance of the proposed CAE-based model is now evaluated using both the standard test set and additional unseen datasets derived from 1D TDSE Lamb wave simulations. As outlined in Table~\ref{tab:Exp_CAE_data_split}, the model was initially trained on 1069 samples and tested on 119 samples from the TDSE dataset. The convergence behaviour of the network during the training and validation stages is illustrated in \Fig\ref{fig:Loss_vs_Epochs_case_2}(a) through the variation of loss with epochs, demonstrating the adequacy of the selected number of training epochs  for the simulated 1D TDSE dataset. 

\begin{figure}[!htb]
    \centering
    \begin{minipage}{0.49\textwidth}
        \centering
        \includegraphics[trim = 0.7in 1.2in 1.2in 0.6in,width=\textwidth]{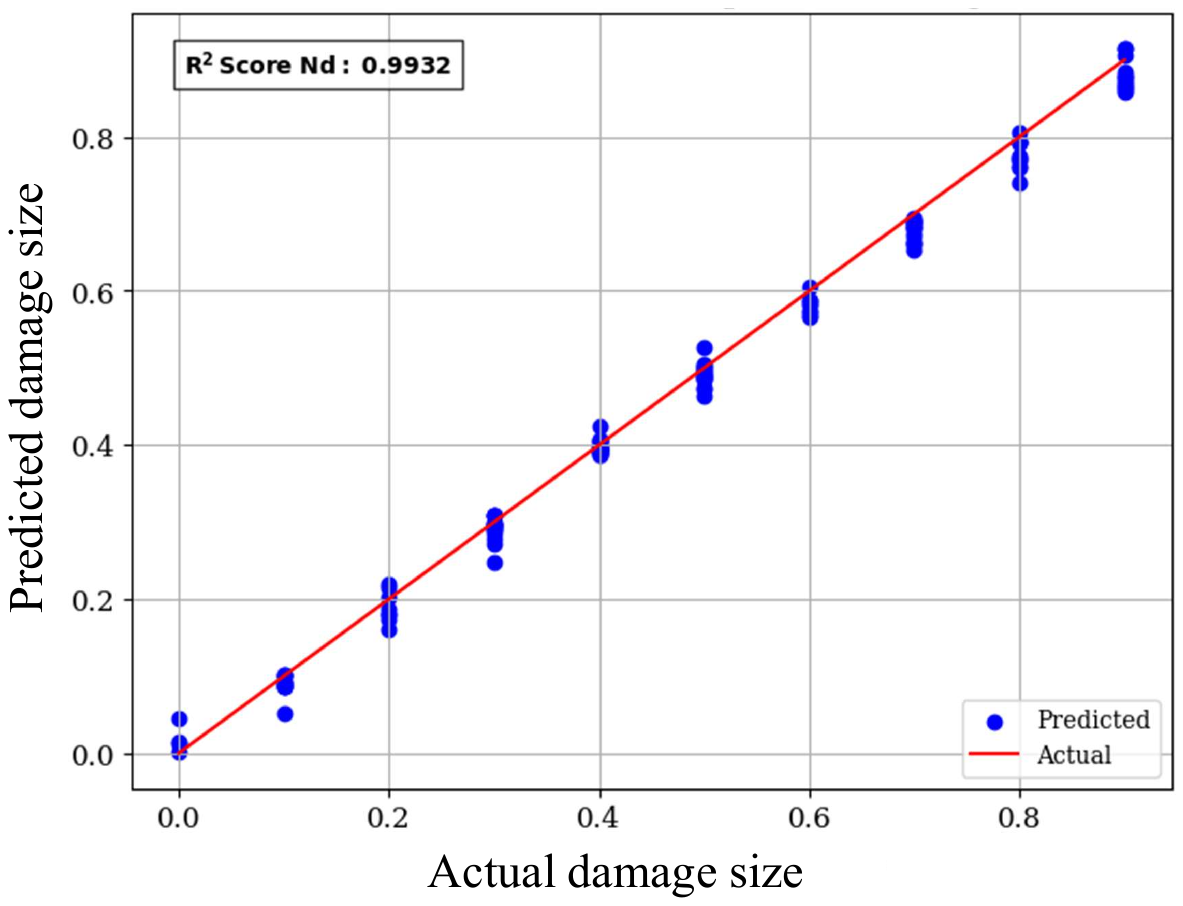}\\
        \small (a)
    \end{minipage}
    \hfill
    \begin{minipage}{0.49\textwidth}
        \centering
        \includegraphics[trim = 0.7in 1.2in 1.2in 0.6in,width=\textwidth]{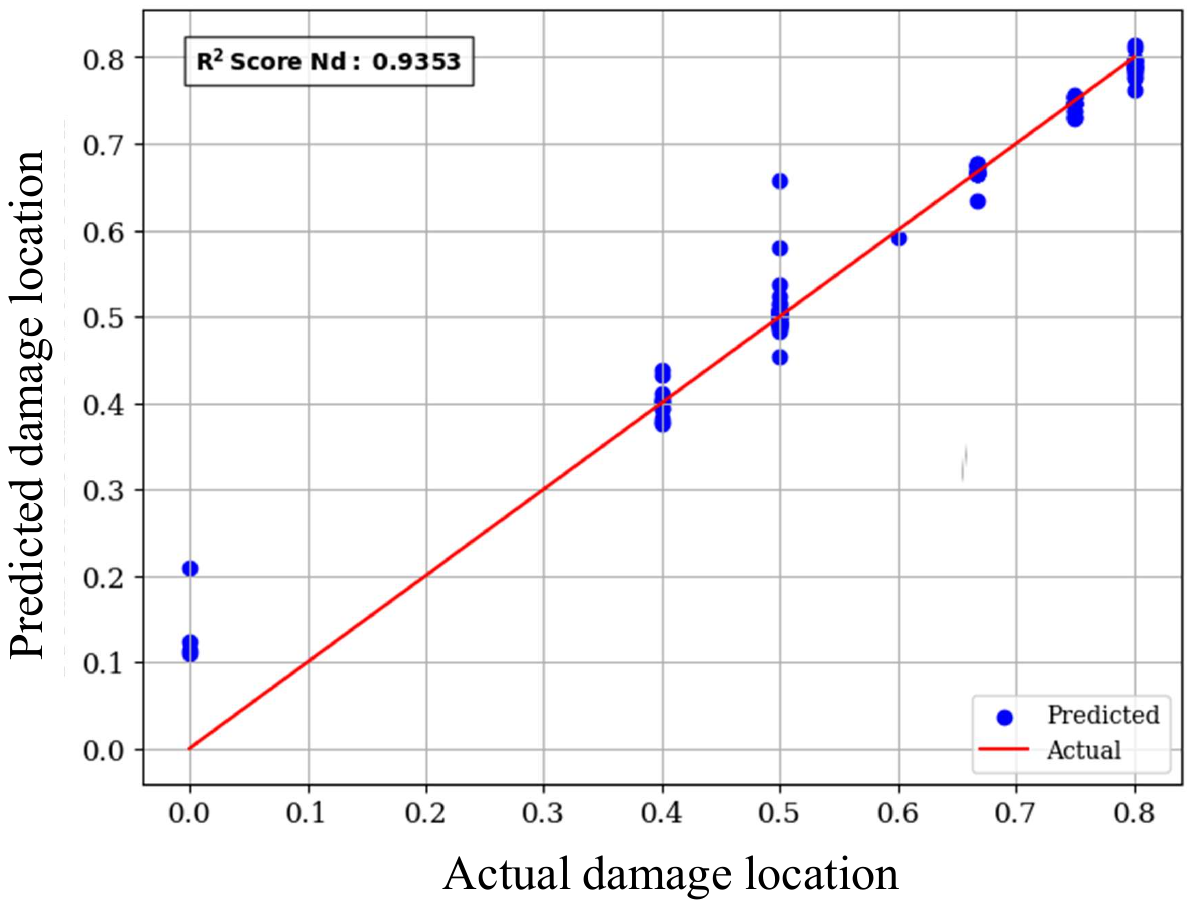}\\
        \small (b)
    \end{minipage}
    \caption{Scatter plots of actual and predicted (a) damage size and (b) location from 1D TDSE test data}
    \label{fig: test_Nd_Np_1}
\end{figure}

Model performance on the primary test set is quantified using the coefficient of determination ($R^2$), evaluating the accuracy of both notch depth and notch location predictions (\Fig\ref{fig: test_Nd_Np_1}). The model achieved $R^2$ scores of 0.9932 for notch depth and 0.9353 for notch location (Table~\ref{tab:CAE_TDSE_Model}), indicating excellent predictive capability on test data drawn from the same distribution as the training set.
\begin{table}[!htb]
\centering
\caption{Performance of CAE-FFNN model trained on 1D TDSE variable-frequency dataset with test and unseen data}
\renewcommand{\arraystretch}{1.6}
\setlength{\tabcolsep}{10pt}
\begin{tabular}{p{5.2cm}ccc}
\hline
{Evaluation case} & {No. of samples} & \multicolumn{2}{c}{{$R^2$ score}} \\
\cline{3-4}
 & & {Notch depth} & {Notch location} \\
\hline
{Test dataset} & 119 & 0.9932 & 0.9353 \\
\hline
{Unseen dataset (unseen notch depth)} & 198 & 0.9852 & 0.9860 \\
\hline
{Unseen dataset (unseen notch location and depth)} & 83 & 0.9520 & 0.8935 \\
\hline
\end{tabular}

\label{tab:CAE_TDSE_Model}
\end{table}

\subsubsection{Evaluation with unseen 1D TDSE data}
\begin{figure}[!htb]
    \centering
    \begin{minipage}{0.49\textwidth}
        \centering
        \includegraphics[trim = 0.9in 1.2in 1.1in 0.6in,width=\textwidth]{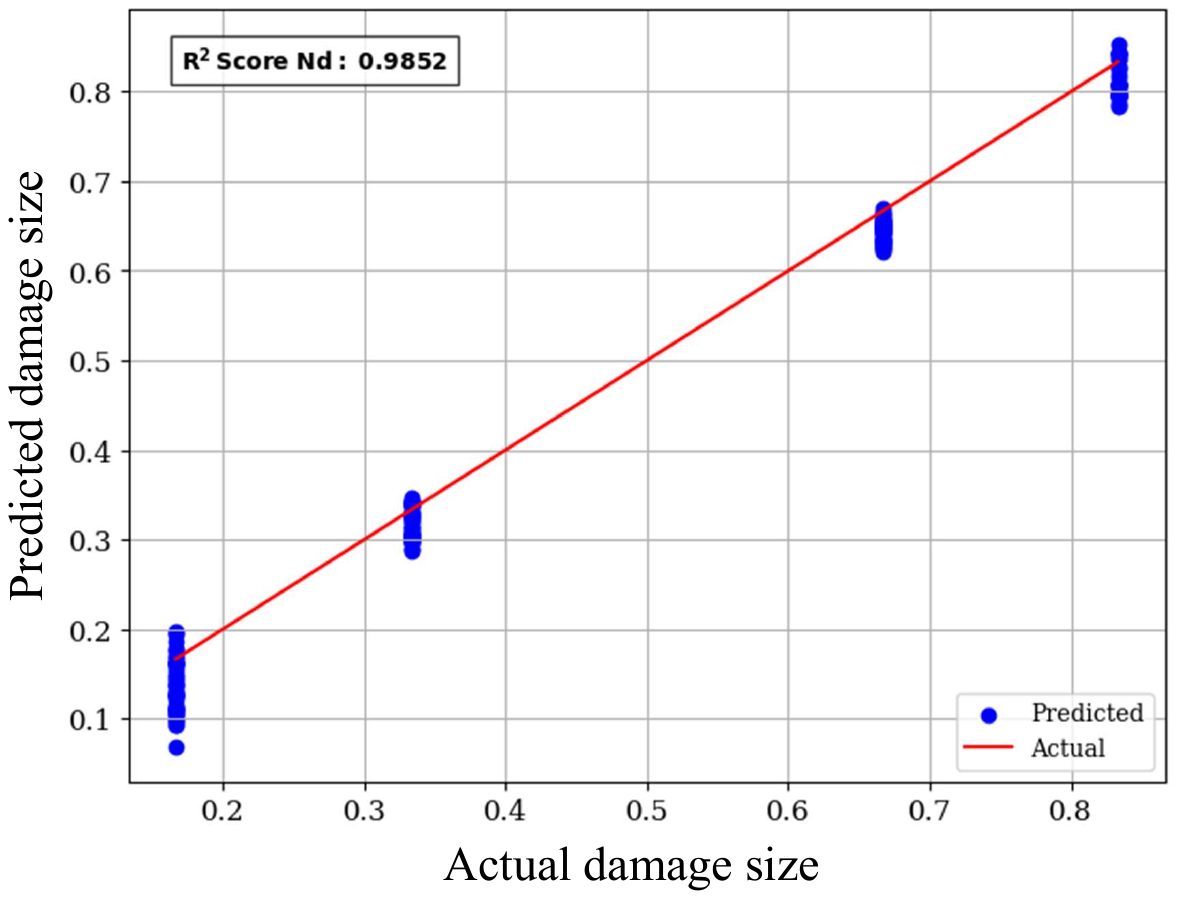}\\
        \small (a)
    \end{minipage}
    \hfill
    \begin{minipage}{0.49\textwidth}
        \centering
        \includegraphics[trim = 0.9in 1.2in 1.1in 0.6in,width=\textwidth]{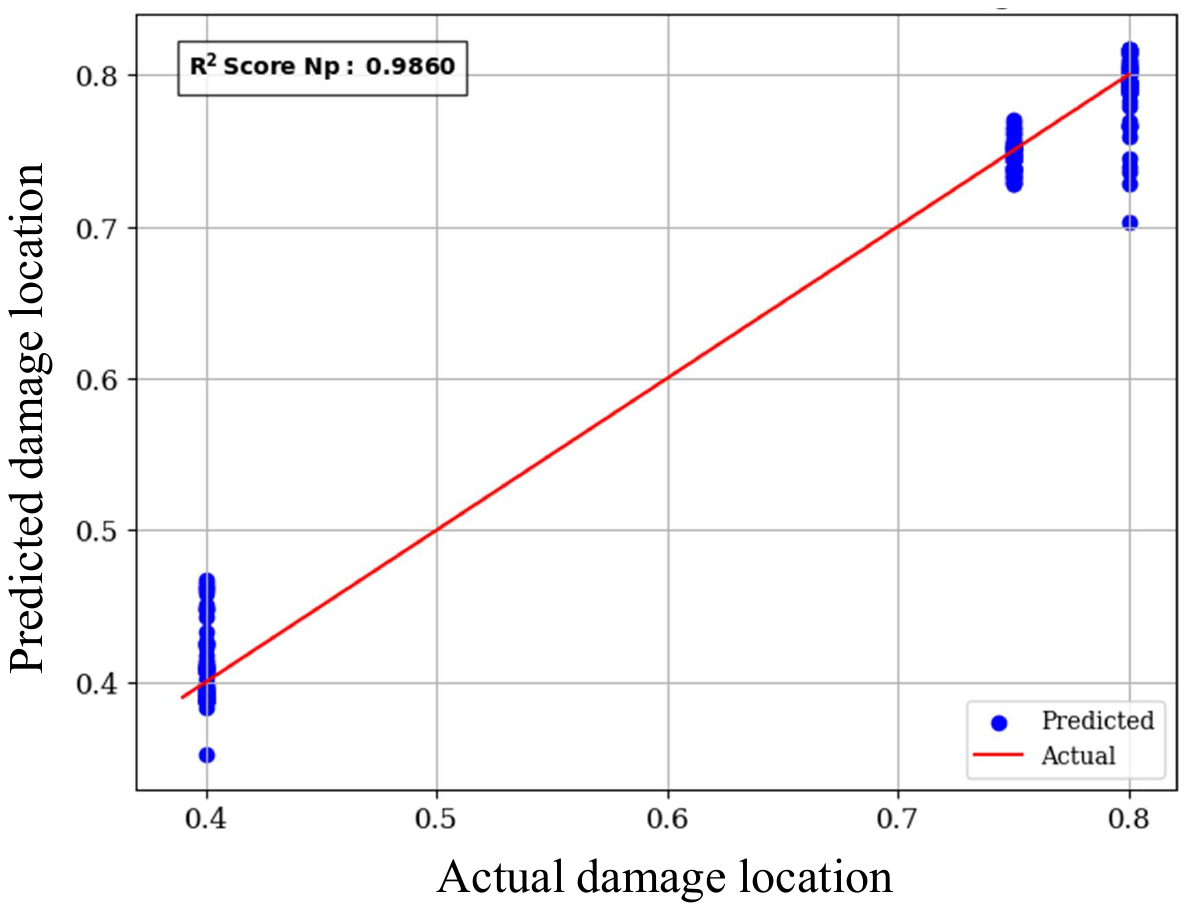}\\
        \small (b)
    \end{minipage}
    \caption{Scatter plots of actual and predicted (a) damage size and (b) location from unseen simulated data}
    \label{fig: TDSE_Nd_Np_unseen}
\end{figure}

To assess the applicability of the model to variations not seen during training, two separate unseen test datasets were used. The first dataset included unseen notch depths while maintaining consistent notch locations and consisted of 198 test samples. The model maintained strong performance, achieving $R^2$ scores of 0.9852 and 0.9860 for depth and location prediction, respectively (\Fig\ref{fig: TDSE_Nd_Np_unseen}).

The second unseen dataset introduced simultaneous variation in both notch depth and location, representing a more challenging evaluation scenario. Despite the increased complexity, the model retained robust performance with $R^2$ scores of 0.9520 for notch depth prediction and 0.8935 for notch location prediction (Table~\ref{tab:CAE_TDSE_Model}). These results demonstrate that the proposed CAE-FFNN model, when trained on a large, noise-augmented TDSE dataset, can accurately predict damage characteristics even for damage cases that do not belong to the training dataset.

\subsubsection{Evaluation with experimental data after transfer learning}
\begin{figure}[!htb]
    \centering
    \begin{minipage}{0.49\textwidth}
        \centering
        \includegraphics[trim = 0.9in 1.2in 1.0in 0.6in,width=\textwidth]{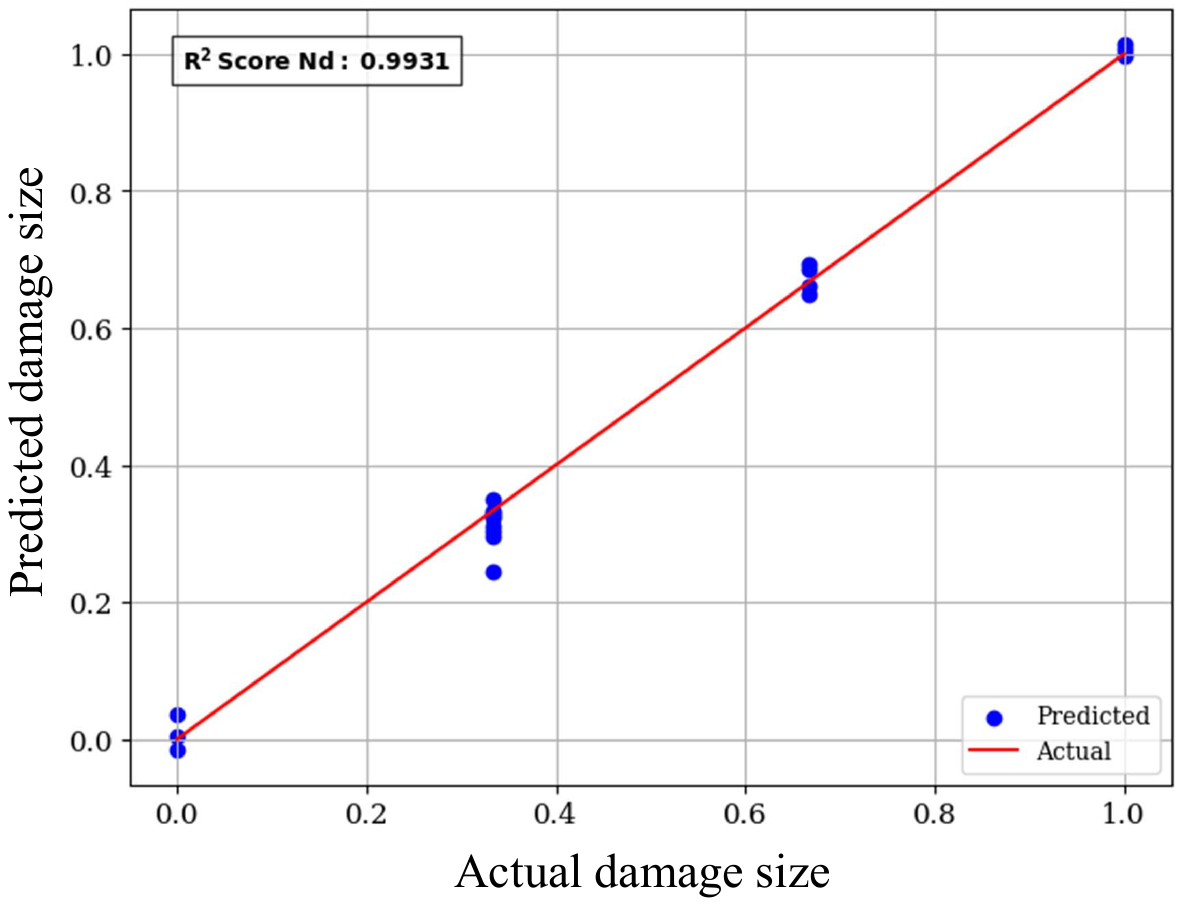}\\
        \small (a)
    \end{minipage}
    \hfill
    \begin{minipage}{0.49\textwidth}
        \centering
        \includegraphics[trim = 0.9in 1.2in 1.0in 0.6in,width=\textwidth]{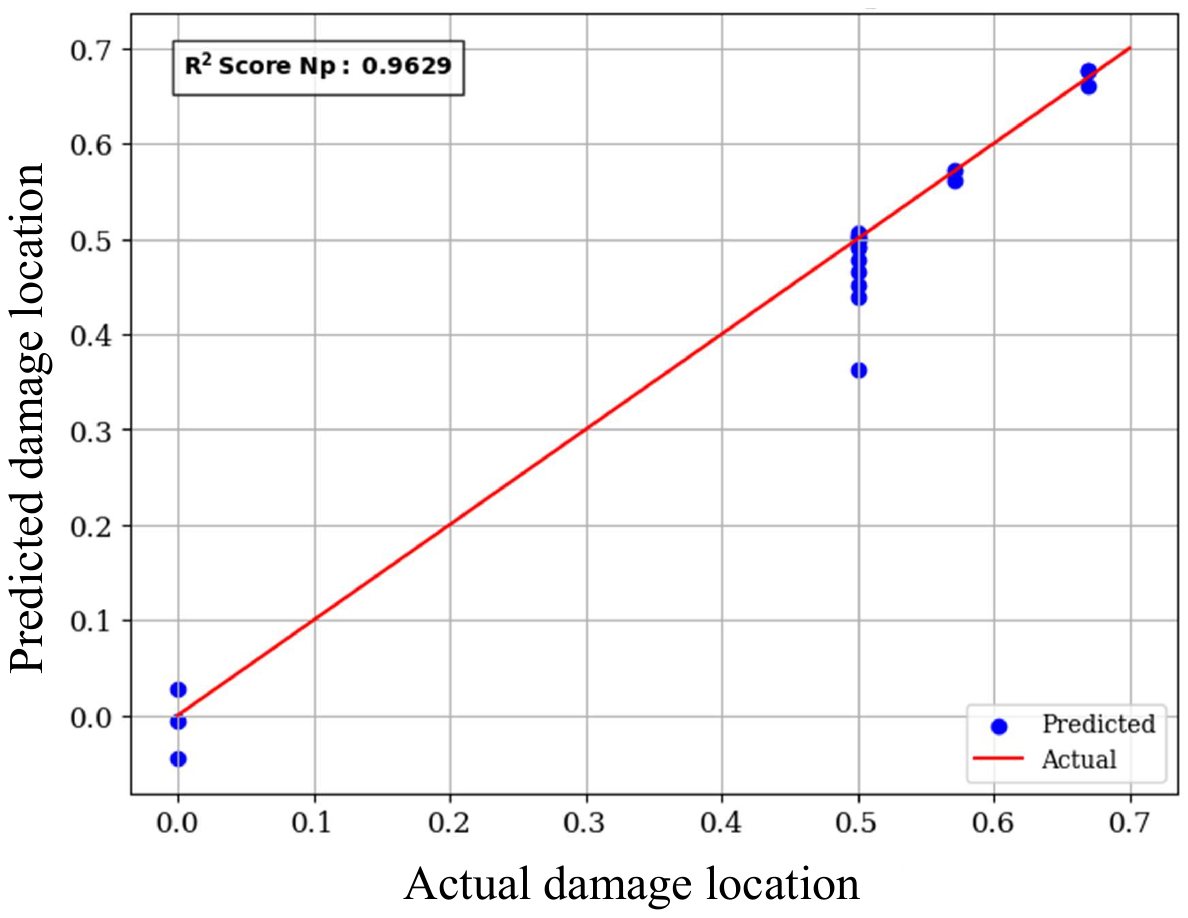}\\
        \small (b)
    \end{minipage}
    \caption{Scatter plots of actual and predicted (a) damage size and (b) location from limited experimental dataset after transfer learning}
    \label{fig: Exp_Nd_Np_test}
\end{figure}

After pretraining and evaluating the model on the large-scale simulated dataset generated with the 1D TDSE, the next step was to fine-tune the pretrained model on a limited set of experimentally acquired Lamb wave signals. Figure~\ref{fig:Loss_vs_Epochs_case_2}(b) demonstrates the adequacy of the selected number of epochs at the fine-tuning stage. Table~\ref{tab:CAE_TDSE_Model_Fine_Tune_Exp} summarises the performance of the fine-tuned model on the test dataset, demonstrating excellent predictive capability with $R^2$ scores of 0.9931 for damage size and 0.9629 for damage location (\Fig\ref{fig: Exp_Nd_Np_test}). 
\begin{table}[!htb]
\centering
\caption{Performance of the proposed CAE-FFNN-TL model after fine-tuning with small experimental dataset}
\renewcommand{\arraystretch}{1.4}
\setlength{\tabcolsep}{10pt}

\begin{tabular}{p{5.2cm}ccc}
\hline
{Evaluation case} & {No. of samples} & \multicolumn{2}{c}{{$R^2$ score}} \\
\cline{3-4}
 & & {Damage size} & {Damage location} \\
\hline
{Experimental test dataset} & 24 & 0.9931 & 0.9629 \\
\hline
{Unseen dataset (unseen damage size and location)} & 26 & 0.9851 & 0.8887 \\
\hline
\end{tabular}
\label{tab:CAE_TDSE_Model_Fine_Tune_Exp}
\end{table}
     
\subsubsection{Evaluation with unseen experimental data}
\begin{figure}[!htb]
    \centering
    \begin{minipage}{0.49\textwidth}
        \centering
        \includegraphics[trim = 0.9in 1.2in 1.1in 0.6in,width=\textwidth]{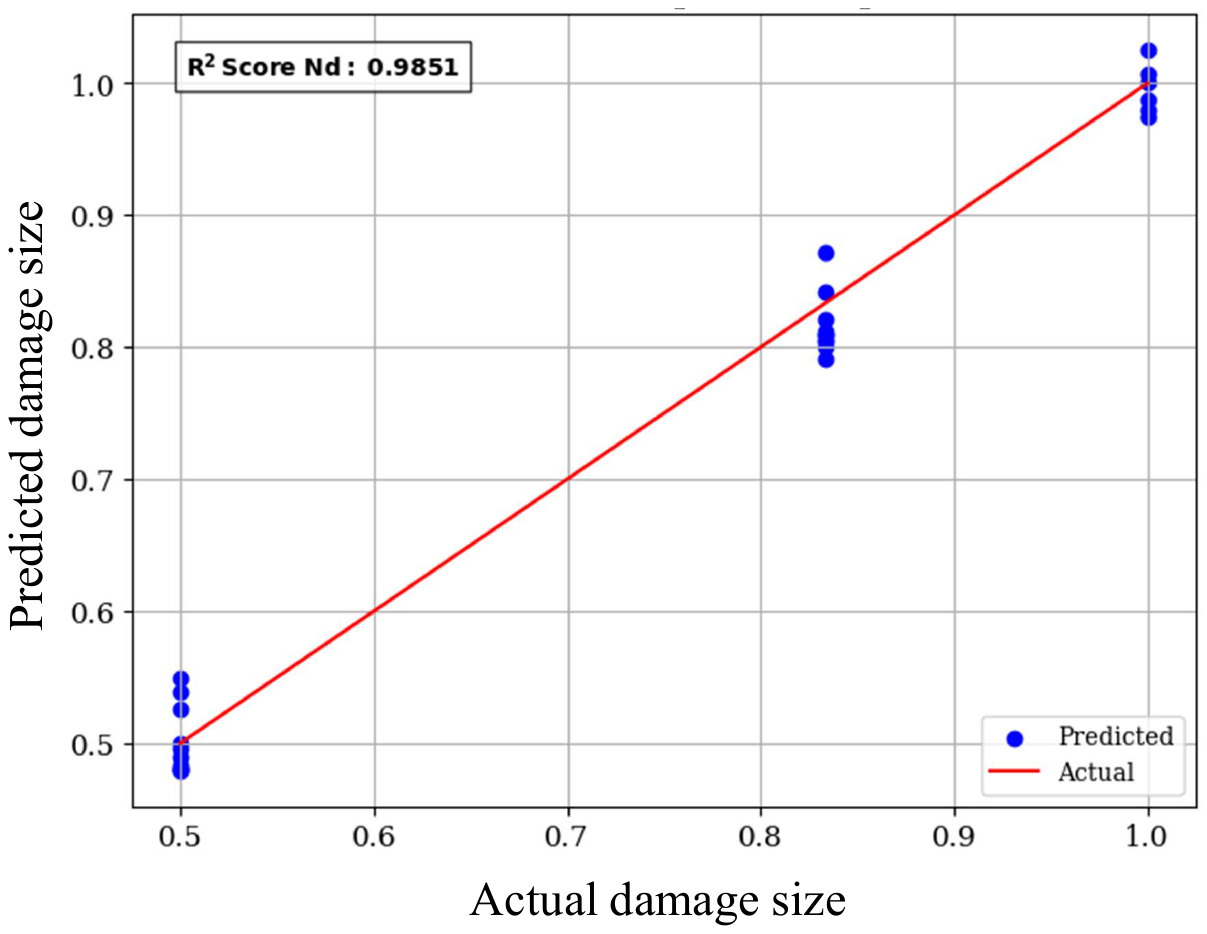}\\
        \small (a)
    \end{minipage}
    \hfill
    \begin{minipage}{0.49\textwidth}
        \centering
        \includegraphics[trim = 0.9in 1.2in 1.1in 0.6in,width=\textwidth]{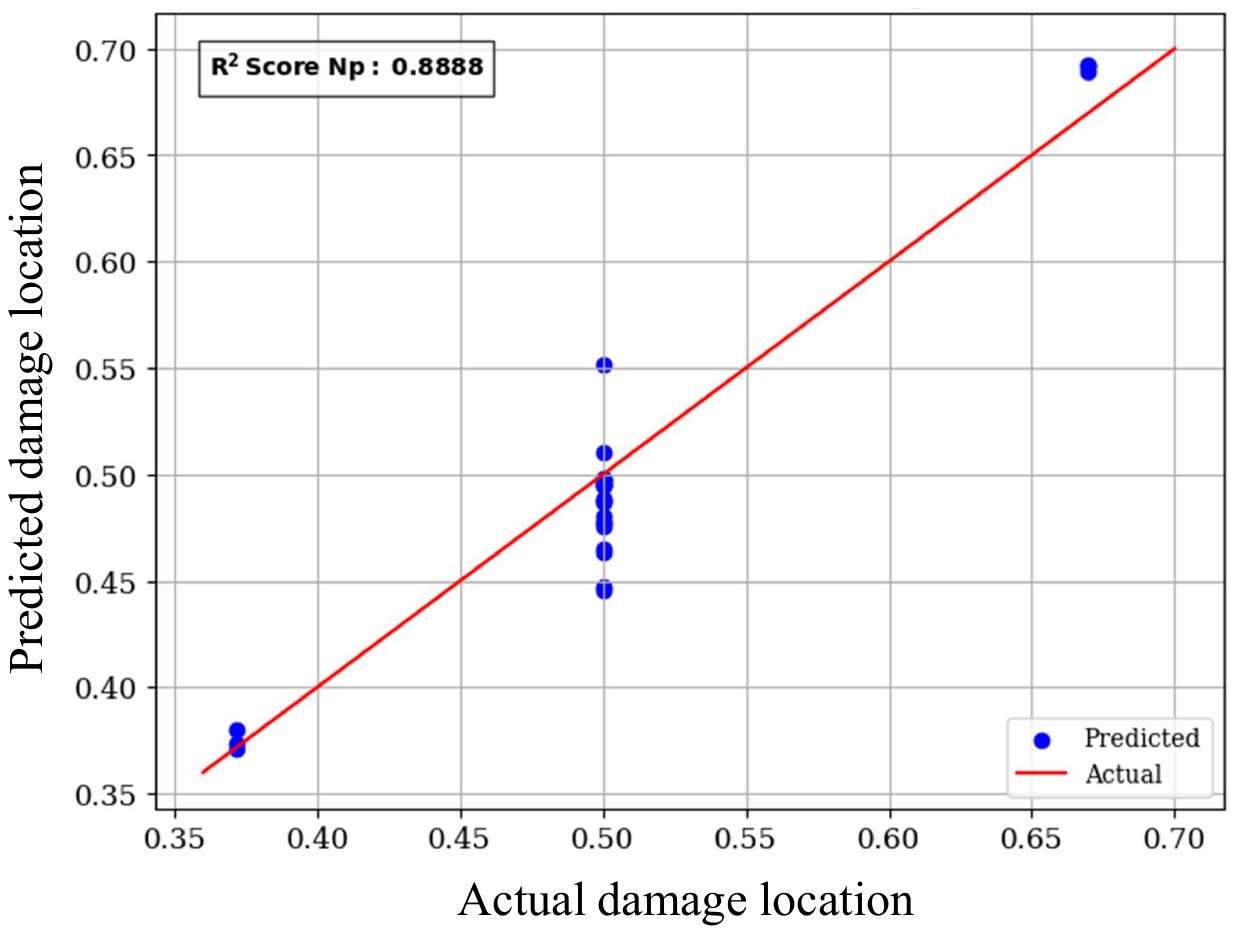}\\
        \small (b)
    \end{minipage}
    \caption{Scatter plots of actual and predicted (a) damage size and (b) location from unseen experimental data after transfer learning}
    \label{fig: Exp_Nd_Np_unseen}
\end{figure}

To further evaluate the robustness and generalisation of the fine-tuned model, it was tested on an unseen experimental dataset comprising 26 samples with variations in both notch location and depth that were not present in the original dataset. The scatter plots of their actual and predicted values presented in \Fig\ref{fig: Exp_Nd_Np_unseen} show that the model maintains strong performance, achieving $R^2$ scores of 0.9851 for notch depth prediction and 0.8887 (Table~\ref{tab:CAE_TDSE_Model_Fine_Tune_Exp}) for notch location prediction. 

These results confirm that the transfer learning approach successfully adapts the model to experimental conditions while maintaining high predictive accuracy despite the limited size of the real-world dataset. Furthermore, it is noteworthy that the block-mass damage considered in the experimental setup is different in nature from the notch-type damage used to generate the simulated dataset for pretraining. Therefore, the results provide strong evidence that the proposed deep TL framework can generalise across distinct damage scenarios through fine-tuning with only a limited amount of corresponding data.

\subsection{Comparison of the proposed CAE-FFNN-TL model and CAE-FFNN model trained on only experimental data}  
Finally, to assess the effectiveness of the proposed strategy of using a large low-cost simulated dataset, a comparative analysis was performed in which the CAE-FFNN model was trained solely on the limited experimental dataset, without leveraging any prior knowledge from the simulated data. The dataset configuration was kept identical, consisting of 168 training signals and 24 testing signals.

\begin{figure}[!htb]
    \centering
    \begin{minipage}{0.49\textwidth}
        \centering
        \includegraphics[trim = 0.9in 1.2in 1.0in 0.6in,width=\textwidth]{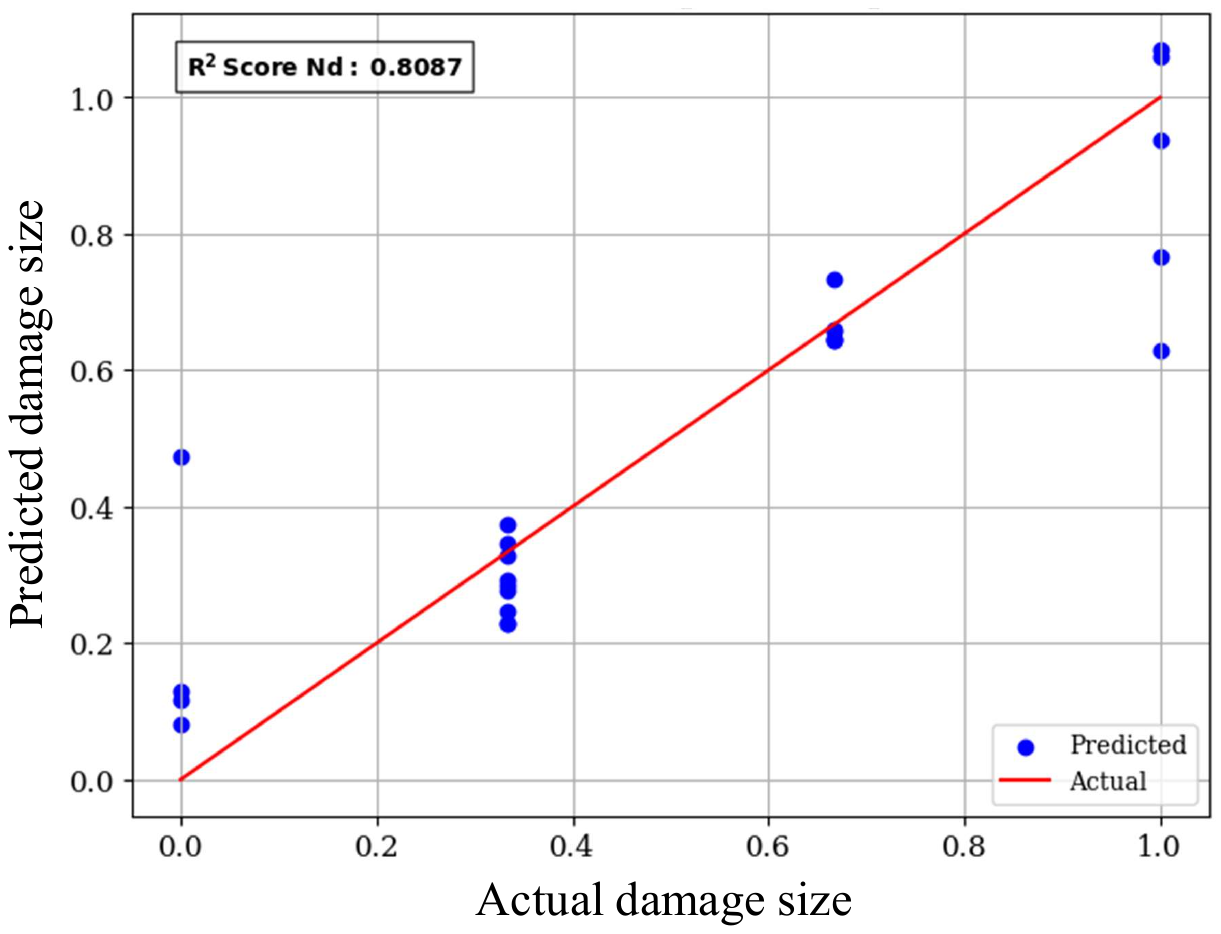}\\
        \small (a)
    \end{minipage}
    \hfill
    \begin{minipage}{0.49\textwidth}
        \centering
        \includegraphics[trim = 0.9in 1.2in 1.0in 0.6in,width=\textwidth]{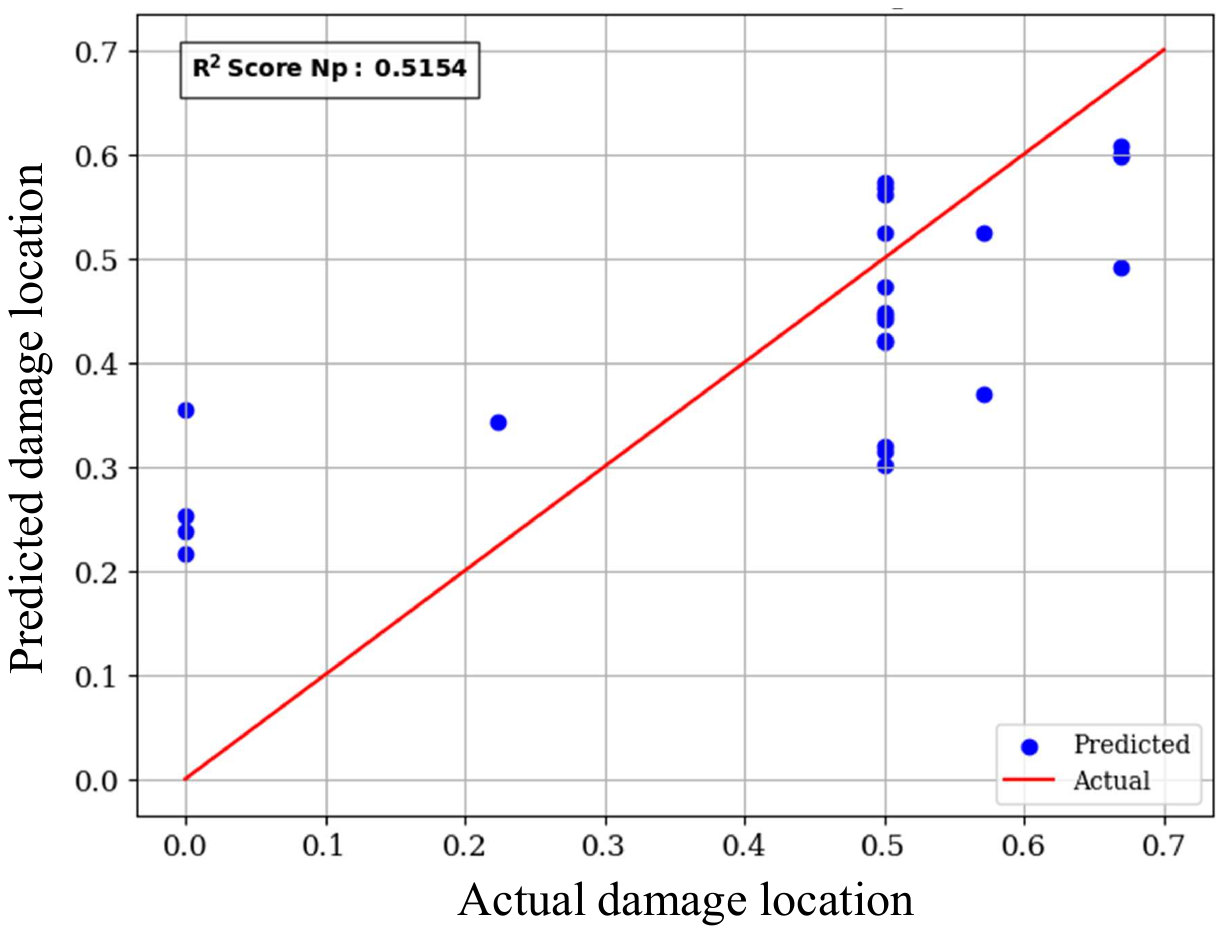}\\
        \small (b)
    \end{minipage}
    \caption{Scatter plots of actual and predicted (a) damage size and (b) location from CAE-FFNN model trained on only limited experimental data}
    \label{fig: Exp_Nd_Np_only}

\end{figure}

The scatter plot presented in \Fig\ref{fig: Exp_Nd_Np_only} for this model show a substantial degradation in predictive performance, yielding $R^2$ values of 0.8087 for notch depth estimation and 0.5154 for notch location prediction. These values are markedly lower than those obtained with the transfer learning-based framework (see Table~\ref{tab:CAE_TDSE_Model_Fine_Tune_Exp}), which achieved $R^2$ scores of 0.9931 and 0.9629 for notch depth and location, respectively. This performance disparity underscores the inherent limitations of training deep learning models solely on small experimental datasets, which typically lack sufficient variability and representativeness to ensure robust generalisation.

In contrast, the transfer learning model, pretrained on a large and diverse 1D TDSE dataset and subsequently fine-tuned with experimental measurements, effectively leveraged the learned feature representations of Lamb wave signals to achieve superior accuracy on both seen and unseen experimental data. These findings clearly demonstrate that the proposed transfer learning strategy not only improves prediction accuracy but also enhances generalisation under realistic experimental conditions. Consequently, pretraining on simulated data followed by fine-tuning on limited experimental datasets constitutes an effective and practical approach for Lamb wave-based SHM when extensive experimental data is unavailable.

\section{Conclusions}
A comprehensive transfer learning framework for Lamb wave-based SHM has been developed by integrating lightweight numerical simulations, deep learning architectures, and limited experimental measurements to achieve accurate damage localisation and quantification in real-world applications. The proposed framework operates directly on raw Lamb-wave signals, thereby avoiding the need for complex signal preprocessing. A computationally efficient strategy for generating large training datasets has been established using the recently developed 1D-TDSE model based on the zigzag theory. The model accurately captures Lamb wave generation, propagation, sensing, and damage interaction while significantly reducing the computational cost compared with conventional continuum-based FE models. The proposed CAE architecture successfully learns compact latent-space representations of raw Lamb wave signals, which provide an effective basis for subsequent damage localisation and severity estimation using an FFNN.

The major conclusions of this study are as follows:

\begin{itemize}

\item Transfer learning from a large 1D-TDSE dataset to a limited high-fidelity 2D FE dataset of raw Lamb wave signals yielded excellent damage localisation and size prediction accuracies, with $R^2$ scores of 0.9305 and 0.9941, respectively. These results demonstrate the ability of the proposed framework to bridge the fidelity gap between lightweight simulations and high-fidelity numerical models.

\item Compared with the proposed CAE-based framework, the CNN-based transfer learning architecture exhibited substantially inferior localisation performance, achieving an $R^2$ score of only 0.3369. This result highlights the superior capability of the CAE-derived latent-space representation in extracting damage-sensitive features from Lamb wave signals.

\item The framework was successfully transferred from the simulation domain to the experimental domain using only 168 experimentally measured signals, representing less than 15\% of the total dataset, for fine-tuning. The resulting model achieved excellent predictive accuracy on the experimental test dataset, with $R^2$ scores of 0.9931 and 0.9629 for damage size and location, respectively.

\item The model demonstrated strong generalisation capability when evaluated on previously unseen numerical and experimental datasets. High prediction accuracies were maintained even when neither the damage size nor the location was represented in the pretraining and fine-tuning datasets, achieving $R^2$ scores of 0.9851 and 0.8887 for damage size and location, respectively.

\item Training solely on the limited experimental dataset resulted in a substantial degradation in performance ($R^2 = 0.8087$ for damage size and $R^2 = 0.5154$ for damage location), whereas the proposed transfer learning framework achieved corresponding $R^2$ scores of 0.9931 and 0.9629. This clearly demonstrates the effectiveness of leveraging large simulation datasets for pretraining when extensive experimental data is unavailable.
\end{itemize}

Overall, the proposed multifidelity transfer learning framework provides a practical, accurate, and computationally efficient pathway for deploying DL-based Lamb wave SHM systems in real-world applications, where acquiring large volumes of labelled experimental data is often prohibitively expensive or infeasible. Future work will focus on extending the proposed framework to multiple damage scenarios, more complex geometries, varying environmental conditions, and experimental studies involving composite materials and full-scale structural components.

\section*{Acknowledgements}
Santosh Kapuria is thankful to Anusandhan National Research Foundation, INDIA, for support through the J.C. Bose National Fellowship (Grant No. JBR/2023/000025) and to Ministry of Ports, Shipping and Waterways for Grant No. 2022-MT-356529. The authors sincerely acknowledge the assistance of Mayank Jain in simulation data generation and Souvik Jana in experimental data generation.

\section*{Conflict of Interest}
The authors have no conflicts of interest to declare.
%\bibliographystyle{jmr2}
%\begin{harvard}
\printcredits

%\bibliographystyle{cas-model1-num-names}
\bibliographystyle{cas-model2-names}
\bibliography{ref_ML}

\end{document}